\documentclass[runningheads]{llncs}
\usepackage{graphicx}

\usepackage{tikz}
\usepackage{comment} 
\usepackage{amsmath,amssymb} 
\usepackage{color}

\usepackage{hyperref}
\usepackage{cite}
\usepackage{bm}
\usepackage{booktabs}

\usepackage{multirow}
\usepackage{xspace}
\makeatletter
\DeclareRobustCommand\onedot{\futurelet\@let@token\@onedot}
\def\@onedot{\ifx\@let@token.\else.\null\fi\xspace}
\def\eg{\emph{e.g}\onedot} 
\def\ie{\emph{i.e}\onedot} 
 
\def\etc{\emph{etc}\onedot}

\def\etal{\emph{et al}\onedot}
\makeatother

\usepackage{caption}

\begin{document}
\pagestyle{headings}
\mainmatter
\def\ECCVSubNumber{5385}  

\title{DOPE: Distillation Of Part Experts for whole-body 3D pose estimation in the wild}

\titlerunning{DOPE: Distillation Of Part Experts for whole-body 3D pose}
\author{Philippe Weinzaepfel \qquad
Romain Br\'egier \qquad
Hadrien Combaluzier \\
Vincent Leroy \qquad
Gr\'egory Rogez }
\authorrunning{P. Weinzaepfel et al.}
%
\institute{NAVER LABS Europe}
\maketitle

\vspace{-0.4cm}
\begin{center}
 \includegraphics[trim=200 80 200 100,clip, width=0.4\linewidth]{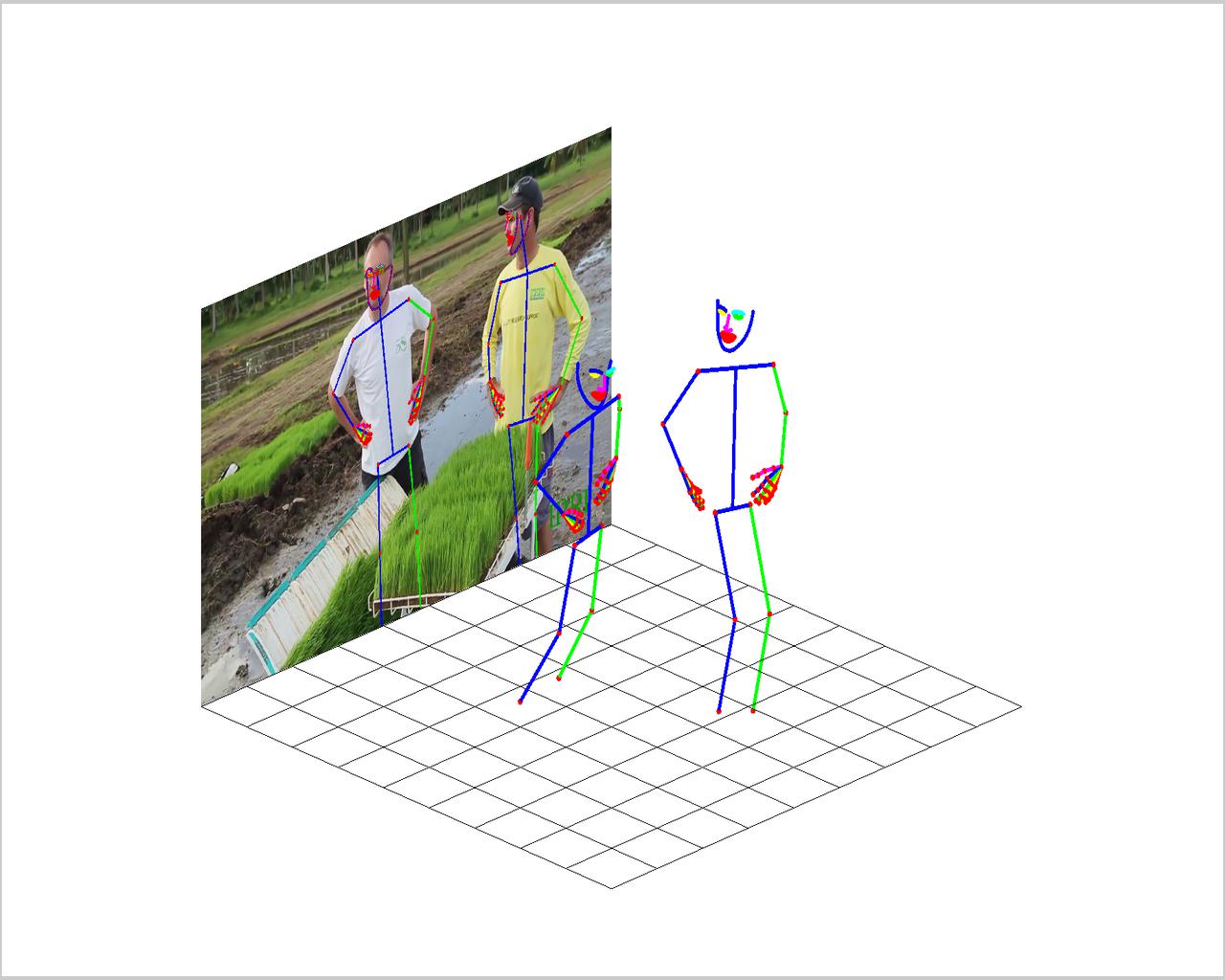}
 \includegraphics[trim=230 30 200 80,clip,width=0.4\linewidth]{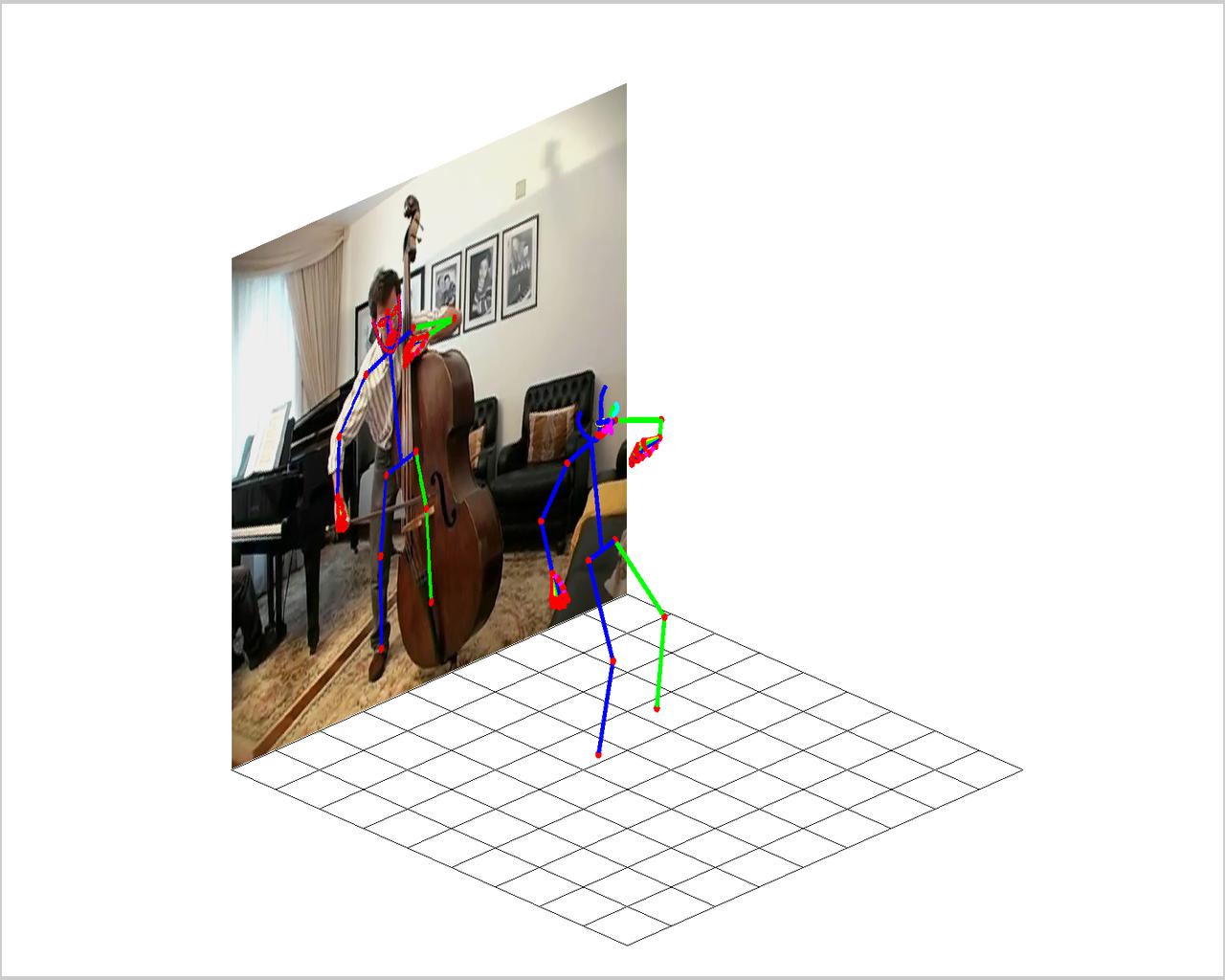}\\[-0.35cm]
 \captionof{figure}{\label{fig:res}  Results of our DOPE approach for 2D-3D whole-body pose estimation.}
 \vspace{-0.1cm}
\end{center}

\begin{abstract}
We introduce DOPE, the first method to detect and estimate whole-body 3D human poses, including bodies, hands and faces, in the wild.
Achieving this level of details is key for a number of applications that require understanding the interactions of the people with each other or with the environment. The main challenge is the lack of in-the-wild data with labeled whole-body 3D poses. In previous work, training data has been annotated or generated for simpler tasks focusing on bodies, hands or faces separately. In this work, we propose to take advantage of these datasets to train independent experts for each part, namely a body, a hand and a face expert, and distill their knowledge into a single deep network designed for whole-body 2D-3D pose detection. In practice, given a training image with partial or no annotation, each part expert detects
its subset of keypoints in 2D and 3D and the resulting estimations are combined to obtain whole-body pseudo ground-truth poses.
A distillation loss encourages the whole-body predictions to mimic the experts' outputs.
Our results show that this approach significantly outperforms the same whole-body model trained without distillation while staying close to the performance of the experts. Importantly, DOPE is computationally less demanding than the ensemble of experts and can achieve real-time performance. 
Test code and models are available at \url{https://europe.naverlabs.com/research/computer-vision/dope}.

\end{abstract}

\section{Introduction}
 
Understanding humans in real-world images and videos has numerous potential applications 
ranging from avatar animation for augmented and virtual reality~\cite{xnect,ARposer} to robotics~\cite{gui2018teaching,garcia2019human}.
To fully analyze the interactions of people with each other or with the environment, and to recognize their emotions or activities, a detailed pose of the whole human body would be beneficial.
This includes 3D body keypoints, \ie, torsos, arms and legs, that give information on the global posture of the persons, but also detailed information about hands and faces to fully capture their expressiveness. The task of whole-body 3D human pose estimation has been mainly addressed part by part as indicated by the large literature on estimating 3D body pose~\cite{lcrnet++,xnect,moon2019camera,up3d,arnab2019exploiting,habibie2019wild}, 3D hand pose~\cite{renderedhand,cai2018,yuan2018,ganerated} or 3D face landmarks and shape~\cite{LS3Dw,Sanyal_2019_CVPR} in the wild. These methods now reach outstanding performances on their specific tasks, and combining them in an efficient way is an open problem. 
 
More recently, a few approaches have been introduced that capture body, hands and face pose jointly.
Hidalgo \etal~\cite{wholebody} extend OpenPose~\cite{openpose}  to predict 2D whole-body poses in natural images. 
To train their multi-task learning approach, they partly rely on datasets for which adding 2D pose annotations is possible, \eg, adding 2D hand pose annotations~\cite{openposehand} to the MPII body pose dataset~\cite{mpii}. Such annotation scheme is not possible when dealing with 3D poses. Importantly, they observe global failure cases when a significant part of the target person is occluded or outside of the image boundaries. Some other works have leveraged expressive parametric human models composed of body, hand and face components stitched together such as Adam~\cite{adam,totalcapture} or SMPL-X~\cite{smplx}.  
These optimization-based approaches remain sensitive to initialization and are usually slow to converge. Their performance highly depends on the intermediate estimation of the 3D orientations of body parts or  2D keypoint locations, and is therefore limited in cases of occlusions or truncations at the image boundary compared to more direct approaches.

In this paper, we propose the first learning-based method that, given an image, detects the people present in the scene and directly predicts the 2D and 3D poses of their bodies, hands and faces, see examples in Figure~\ref{fig:res}.
Inspired by LCR-Net++~\cite{lcrnet++}, a Faster R-CNN like architecture~\cite{Faster} tailored for in-the-wild 2D-3D body pose estimation, we design a classification-regression network where the object categories to detect are body, hand and face pose classes. In a second step, a class-specific regression is applied to refine body, hand and face pose estimates by deforming the average pose of each class both in 2D and 3D.

There exists no in-the-wild dataset to directly train our network, \ie, images
with 3D pose annotations for body, hand and face poses. Such data could only be obtained in specific controlled environments, \eg in motion capture rooms or through computer-generation, which would not suit our purpose of whole-body pose estimation in unconstrained scenarios. However, multiple in-the-wild datasets are available for each independent task, \ie, for 3D body pose estimation~\cite{xnect,up3d}, 3D hand pose estimation~\cite{stereohand,renderedhand,honnotate} or 3D facial landmark estimation~\cite{LS3Dw,menpo}. 
Task-specific methods trained on these datasets perform well in practice but our experiments show that training our single model for whole-body 3D pose estimation on the union of these datasets leads to poor performances. 
Each dataset being annotated with partial pose information (\ie, its specific part), unannotated parts are mistakenly considered as negatives by our detection framework, burdening  the performance of the network.

\begin{figure}[t]
 \centering
 \includegraphics[width=\linewidth]{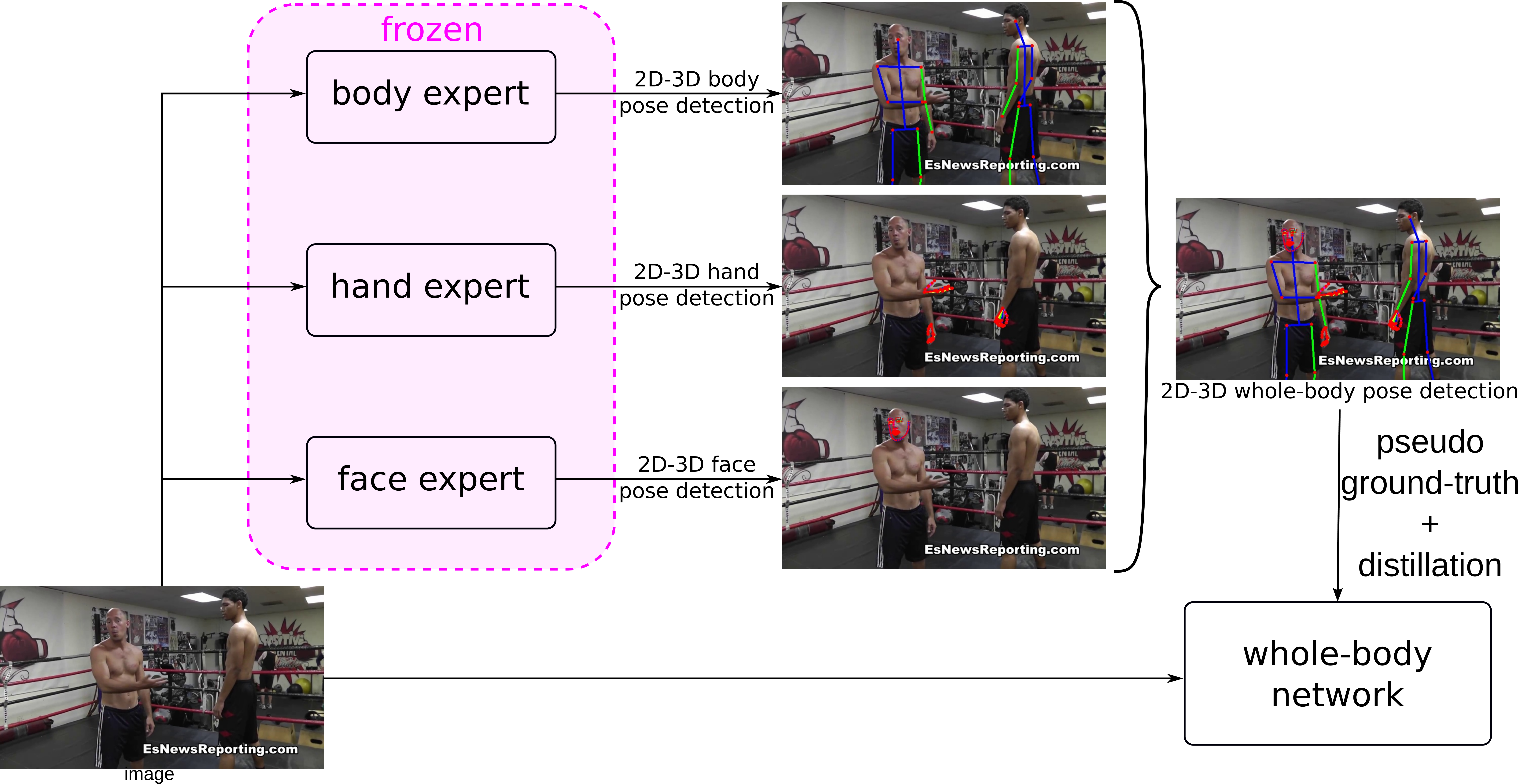}\\[-0.3cm]
 \caption{Overview of our DOPE training scheme. Each training image is processed by the part experts to detect their specific parts and estimate their 2D and 3D poses. The resulting detections are combined to obtain the whole-body poses used as ground-truth for this image when training our network. 
 We show only 2D poses for the sake of clarity but we also distill the 3D poses.}
 \vspace{-0.2cm}
 \label{fig:train}
\end{figure}

To handle this problem, we propose to train independent experts for each part, namely body, hand and face experts, and distill their knowledge to our whole-body pose detection network designed to perform the three tasks jointly. 
In practice, given a training image with partial or no annotation, each part expert detects and estimates its subset of keypoints, in 2D and 3D, and the resulting estimations are combined to obtain whole-body pseudo ground-truth poses for the whole-body network.
Figure~\ref{fig:train} illustrates this training scheme.
A distillation loss is applied on the network's output to keep it close to the experts' predictions.
We name our method \textbf{DOPE} for \textbf{D}istillation \textbf{O}f \textbf{P}art \textbf{E}xperts.
Our unified DOPE model performs on par with the part experts when evaluating each of the three tasks on dedicated datasets, while being computationally less demanding than the ensemble of experts and achieving real-time performances. In summary, we propose (a) a new architecture that can detect and estimate the whole-body 2D-3D pose of multiple people in the wild  in real-time and (b) a novel and effective training scheme based on distillation that leverages previous data collection efforts for the individual subtasks. 

This paper is organized as follows. 
After reviewing the related work (Section~\ref{sec:related}), we present our DOPE method in Section~\ref{sec:dope}.
Finally, experimental results for body, hand and face pose estimation are reported in Section~\ref{sec:xp}.

\section{Related work}
\label{sec:related}

The problem of 3D human whole-body pose estimation has been mainly tackled by breaking the body into parts and focusing on the pose inference of these parts separately. In the following, we briefly review the state of the art for each  of these subtasks, before summarizing the few approaches that predict the 3D pose of the entire body, and finally discussing existing distillation methods.

\noindent \textbf{3D body pose estimation.}
Two basic categories of work can be found in the recent literature: (a) approaches that directly estimate the 3D body keypoints from an input image~\cite{RogezS16,PavlakosZDD17a,lcrnet++,xnect,moon2019camera} and (b) methods that leverage 2D human pose estimation~\cite{MartinezHRL17,ChenR17,BogoKLG0B16,up3d}. 
The latter ones rely on a previous localization of the body keypoints in the image, through an off-the-shelf 2D pose detector~\cite{openpose,alphapose,maskrcnn}, and lift them to 3D space~\cite{MartinezHRL17,ChenR17} or, as in \cite{BogoKLG0B16,up3d}, use them to initialize the optimization procedure of a parametric model of the human body such as SMPL~\cite{smpl}. For our body expert, we employ LCR-Net++~\cite{lcrnet++} that jointly estimates 2D and 3D body poses from the image and has demonstrated robustness to challenging in-the-wild scenarios, \ie, showing multiple interacting persons, with cluttered backgrounds and/or captured under severe occlusions and truncations.

\noindent \textbf{3D hand pose estimation.}
3D hand pose estimation from depth data has been studied for many years and state-of-the-art results on this task are now impressive as shown in a recent survey~\cite{supancic2018}. 
RGB-based 3D hand pose estimation is more challenging, and has gained interest in recent years. 
Regression-based techniques try to directly predict 3D location of hand keypoints~\cite{yang2019} or even the vertices of a mesh~\cite{ge2019} from an input image. 
Some methods incorporate priors by regressing parameters of a deformable hand model such as MANO~\cite{romero2017,boukhayma2019,hasson2019},
and many techniques leverage intermediate representations such as 2D keypoints heatmaps to perform 3D predictions~\cite{renderedhand,cai2018,ganerated,zhang2019}.
However, pose estimation is often performed on an image cropped around a single hand, and hand detection is performed independently. For our hand expert, we therefore use the detector of~\cite{lcrnet++} (adapted to hands) that recently achieved outstanding performances in RGB-based 3D hand pose estimation under hand-object interaction~\cite{ICCV19HandChallenge}.

\noindent \textbf{3D face pose estimation.}
As with hands, the recovery of the pose of a face is typically performed from an image crop, by detecting particular 2D facial landmarks~\cite{wu2019}. To better perceive the 3D pose and shape of a face, some works propose to fit a 3D Morphable Model~\cite{face3dmm_blanz, face3dmm_booth, zhu2017} or to regress dense 3D face representations~\cite{pix2face, jackson2017, feng2018}. In this work, we also adopt~\cite{lcrnet++} as face expert, resulting in an hybrid model-free approach that regresses 3D facial landmarks independently from their visibility, as in the approach introduced for the Menpo 3D benchmark~\cite{menpo}.

\noindent \textbf{3D Whole-body pose estimation.}
The few existing methods~\cite{totalcapture,smplx} that predict the 3D pose of the whole-body all rely on parametric models of the human body, namely Adam~\cite{adam} and SMPL-X~\cite{smplx}. These models are obtained by combining body, hand and face parametric models. Adam stitches together three different models: a simpler version of SMPL for the body, an artist-created rig for the hands, and the FaceWarehouse model~\cite{FaceWarehouse} for the face. In the case of SMPL-X, the SMPL body model is augmented with the FLAME head model~\cite{Flame} and MANO~\cite{romero2017}. A more realistic model is obtained in the case of SMPL-X by learning the shape and pose-dependent blend shapes. Both methods are based on an optimization scheme guided by 2D joint locations or 3D part orientations. Monocular Total Capture~\cite{totalcapture} remains limited to a single person while for SMPL-X~\cite{smplx}, the optimization strategy is applied independently on each person detected by OpenPose~\cite{openpose}. Optimizing over the parameters of such models can be time-consuming and the performance often depends on a correct initialization. Our approach is the first one that predicts whole-body 3D pose without relying on the optimization of a parametric model and can make real-time predictions of multiple 3D whole-body poses in real-world scenes. In addition, our DOPE training scheme can leverage datasets that do not contain ground-truth for all the parts at once.

\noindent \textbf{Distillation.}
Our learning procedure is based on the concept of distillation which was proposed in the context of efficient neural network computation by using class probabilities of a higher-capacity model as soft targets of a smaller and faster model~\cite{hinton2015distilling}. 
Distillation has been successfully employed for several problems in computer vision such as object detection~\cite{chen2017learning}, video classification~\cite{BhardwajSK19}, action recognition~\cite{CrastoWAS19}, multi-task learning~\cite{liu2019improving} or lifelong learning~\cite{Hou2018lifelong}. 
In addition to training a compact model~\cite{chen2017learning,BhardwajSK19}, 
several works \cite{CrastoWAS19,HoffmanHallucination} have shown that distillation can be combined with privileged information~\cite{VapnikTransferLearning}, also called generalized distillation~\cite{lopez2015unifying} in order to train a network while leveraging extra modalities available for training, \eg training on RGB and depth data while only RGB is available at test time. 
In this paper, we propose to use distillation in order to transfer the knowledge of several body-part experts into a unified network that outputs a more complete representation of the whole human body.

\section{DOPE for 2D-3D whole-body pose estimation}
\label{sec:dope}

After introducing our architecture for multi-person whole-body pose estimation (Section ~\ref{sub:archi}), we detail our training procedure based on distillation  (Section~\ref{sub:dope}).

\begin{figure}
 \centering
 \includegraphics[width=\linewidth]{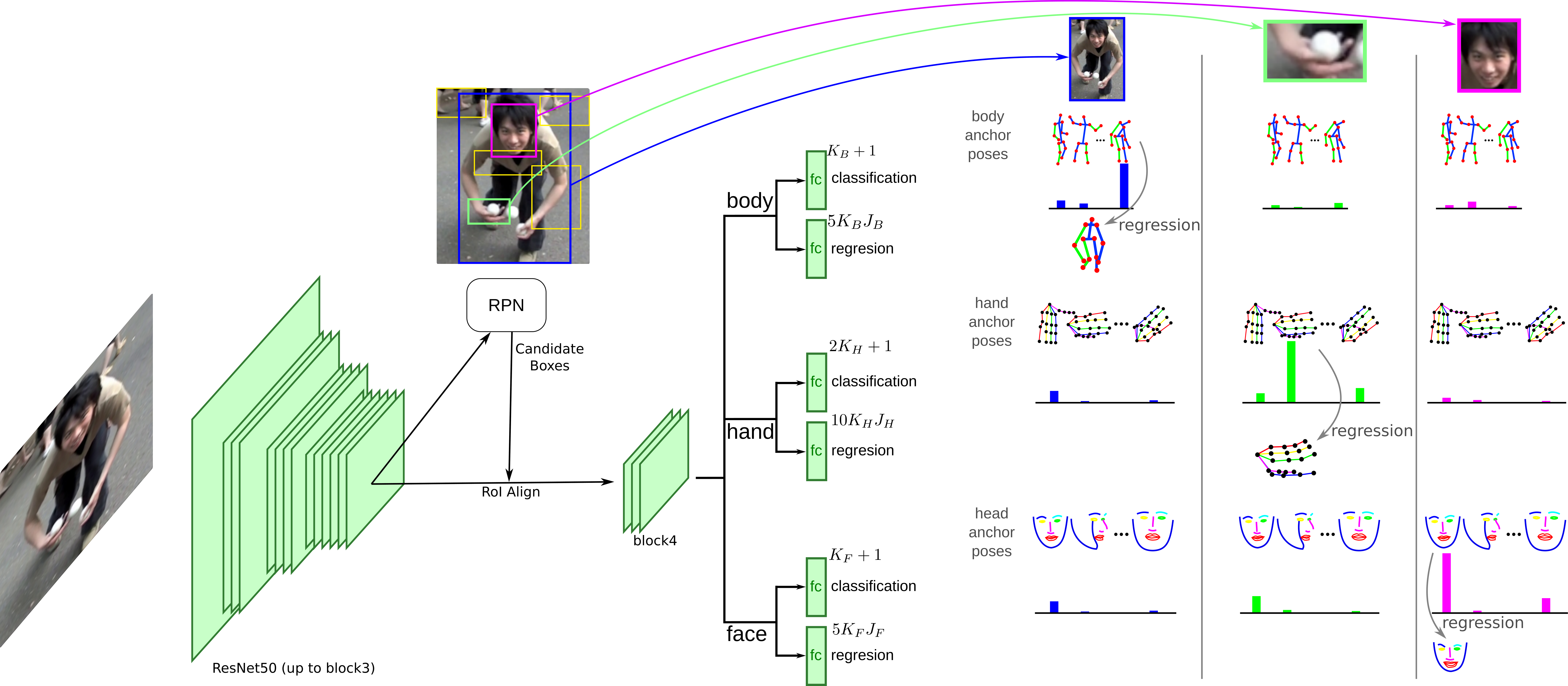}
 \caption{Overview of our whole-body pose estimation architecture. Given an input image, convolutional features are computed and fed into a Region Proposal Network (RPN) to produce a list of candidate boxes. For each box, after RoI-Align and a few additional layers, 6 final outputs are computed (2 for each part). The first one returns a classification score for each anchor-pose corresponding to this part (including a background class not represented for clarity) while the second one returns refined 2D-3D pose estimates obtained through class-specific regression from the fitted anchor pose.}
 \label{fig:archi}
\end{figure}

\subsection{Whole-body pose architecture}
\label{sub:archi}

We propose a method that, given an image, detects the people present in the scene and directly predicts the 2D and 3D poses of their bodies, hands and faces.
Our network architecture takes inspiration from~\cite{lcrnet++}, which extends a Faster R-CNN like architecture~\cite{Faster} to the problem of 2D-3D body pose estimation and has shown to be robust in the wild. 
We thus  design a Localization-Classification-Regression network where the objects to be detected are bodies, hands and faces with respectively $J_B$, $J_H$ and $J_F$ keypoints to be estimated in 2D and 3D. Figure~\ref{fig:archi} shows an overview of this architecture.

\noindent \textbf{Localization.} Given an input image, convolutional features (ResNet50~\cite{resnet} up to block3 in practice) are computed and fed into a Region Proposal Network (RPN) ~\cite{Faster} to produce a list of candidate boxes containing potential body, hand or face instances. Although they might belong to the same person, we specifically treat the parts as separate objects to be robust to cases where only a face, a hand or a body is visible in the image. Our network can also output whole-body poses of multiple people at once, when their different parts are visible.
The candidate boxes generated by the RPN are used to pool convolutional features using RoI Align, and after a few additional layers (block4 from ResNet50 in practice), they are fed to the classification and regression branches, 6 in total: one classification and one regression branch per part.

\noindent \textbf{Classification.} Classification is performed for the three sub-tasks: body, hand and face classification. As in~\cite{lcrnet++}, pose classes are defined by clustering the 3D pose space. This clustering is applied independently in the 3 pose spaces, corresponding to the 3 parts, obtaining respectively a set of $K_B$, $K_H$ and $K_F$ classes for bodies, hands and faces. Note that to handle left and right hands with the same detector, we actually consider $2 \times K_H$ hand classes, $K_H$ for each side. For each classification branch, we also consider an additional background class to use the classifier as a detector. Therefore, each candidate box is classified into $K_B+1$ labels for body classes, $2K_H+1$ for hands and $K_F+1$ for faces.

\noindent \textbf{Regression.}  In a third step, a class-specific regression is applied to estimate body, hand and face poses in 2D and 3D. First, for each class of each part, we define offline the `anchor-poses', computed as the average 2D and 3D poses over all elements in the corresponding cluster. After fitting all the 2D anchor-poses into each of the candidate boxes, we perform class-specific regressions to deform these anchor-poses and match the actual 2D and 3D pose in each box. This operation  is carried out for the 3 types of parts, obtaining $5 \times J_B \times K_B$ outputs for the body part, $5 \times 2 \times J_H \times K_H$  for the hands and $5 \times J_F \times K_F$ for the face. The number $5$ corresponds to the number of dimensions, \ie, 2D+3D.

\noindent \textbf{Postprocessing.} 
For each body, hand or face, multiple proposals can overlap and produce valid predictions. As in~\cite{lcrnet++}, these pose candidates are combined, taking into account their 2D overlap, 3D similarity and classification scores.
To obtain whole-body poses from the independent part detections produced by our network, we simply attach a hand to a body if their respective wrist estimations are close enough in 2D, and similarly for the face with the head body keypoint.

\subsection{Distillation of part experts}
\label{sub:dope}

Even if in-the-wild datasets with 3D pose annotations have been produced for bodies, hands and faces separately, there exists no dataset covering the whole-body at once. 
One possibility is to employ a union of these datasets to train our whole-body model. Since the datasets specifically designed for pose estimation of one part do not contain annotations for the others, \eg body datasets do not have hand and face annotations and vice-versa, unannotated parts are therefore considered as negatives for their true classes in our detection architecture. In practice, this deteriorates the detector's ability to detect these parts and leads to worse overall performances ($\sim$10\% drop for hands and faces, and $\sim$2\% for bodies). To leverage the multiple part-specific datasets, we therefore propose to train independent experts for each part, namely body, hand and face experts, and distill their knowledge into our whole-body pose 
network designed to perform the three tasks jointly.

\noindent \textbf{Part experts.} To ease the distillation of the knowledge, 
we select our 3 experts to match the structure of the classification-regression branches of our whole-body pose estimation architecture and consider the same anchor poses as for the individual tasks. 
We therefore selected the Localization-Classification-Regression network from LCR-Net++~\cite{lcrnet++} as body expert and estimate $J_B=13$ body joints with $K_B=10$ classes.  We  also used the hand detection version of this architecture~\cite{ICCV19HandChallenge}, replacing the $K_B$ body pose classes by $K_H=5$ hand anchor-poses for each side and using the standard number of $J_H=21$ hand  joints: 1 keypoint for the wrist plus 4 for each finger. Finally, to obtain our face expert, we adapted the same architecture to detect 2D-3D facial landmarks. We used the 84 landmarks defined in the 3D Face Tracking Menpo benchmark~\cite{menpo3d} that include eyes, eyebrows, nose, lips and facial contours. We defined $K_F=10$ anchor-poses by applying K-means on all faces from the training set. 
 
\noindent \textbf{Training via distillation.} 
We propose to distill the knowledge of our three part experts to our whole-body pose detection model. 
Let $\mathcal{B}$, $\mathcal{H}$ and $\mathcal{F}$ be the training datasets used for the three individuals tasks, \ie, body, hand, and face pose detection, respectively.
They are associated with ground-truth (2D and 3D) pose annotations for bodies $b$, hands $h$ and faces $f$, respectively. 
In other words, the body expert is for instance trained on $\mathcal{B}=\{ I_i, b_i \}_i$, \ie, a set of images $I_i$ with body ground-truth annotations $b_i$, and similarly for the other parts.

To train our network, we need ground-truth annotations $w$ for the whole body.
We propose to leverage the detections made by the experts in order to augment the annotations of the part-specific datasets.
We denote by $\hat{b}_i$, $\hat{h}_i$ and $\hat{f}_i$ the detections obtained when processing the images $I_i$ with our expert for body, hands and face respectively.
We train our DOPE network on:
\begin{equation}
\mathcal{W}_{DOPE}= \{I_i, w_i\}_{i \in \mathcal{B} \cup \mathcal{H}  \cup \mathcal{F}} \mbox{~~~~where~~} w_i= \left\{
    \begin{array}{ll}
       \{b_i, \hat{h}_i, \hat{f}_i \}  & \mbox{if }    i \in  \mathcal{B}~,  \\
       \{ \hat{b}_i,h_i, \hat{f}_i \}  & \mbox{if }   i \in  \mathcal{H}~, \\
       \{ \hat{b}_i, \hat{h}_i, f_i\}  & \mbox{if }   i \in  \mathcal{F}~.\\
    \end{array}
\right.
\end{equation}
The detections $\hat{b}_i$,  $\hat{h}_i$ and $\hat{f}_i$ estimated by the experts are therefore considered as pseudo ground-truth for the missing keypoints in 2D and 3D. 
In practice, ground-truth annotations are completed using these estimations, for example when some annotations have been incorrectly labeled or are simply missing. 
Note that training images with no annotation at all could also be used to train our network, using only pseudo ground-truth annotations~\cite{lee2013pseudo}, \ie, $w_i=  \{ \hat{b}_i, \hat{h}_i, \hat{f}_i \}$. The training scheme is illustrated in Figure~\ref{fig:train}.  

\noindent \textbf{Loss.}
Our loss $\mathcal{L}$ to train the network combines the RPN loss $\mathcal{L}_{RPN}$ as well as the sum of three terms for each part $p \in \{\text{body,hand,face}\}$:
(a) a classification loss $\mathcal{L}^{p}_{cls}$, (b) a regression loss $\mathcal{L}^{p}_{reg}$, (c) a distillation loss $\mathcal{L}^{p}_{dist}$:
\begin{equation}
 \mathcal{L} = \mathcal{L}_{RPN} + \sum_{p \in \{\text{body,hand,face}\}} \mathcal{L}^{p}_{cls} + \mathcal{L}^{p}_{reg} + \mathcal{L}^{p}_{dist}~~,
\end{equation}
\noindent where $\mathcal{L}_{RPN}$  is the RPN loss from Faster R-CNN~\cite{Faster}.
The classification loss  $\mathcal{L}^{p}_{cls}$ for each part $p$ is a standard softmax averaged over all boxes.
 If a box sufficiently overlaps with a ground-truth box, its ground-truth label is obtained by finding the closest anchor-pose from the ground-truth pose. Otherwise it is assigned a background label, \ie, 0.

The regression loss $\mathcal{L}^{p}_{reg}$ is a standard L1 loss on the offset between  ground-truth 2D-3D poses and their ground-truth anchor-poses, averaged over all boxes.
Note that the regression is class-specific, and the loss is only applied on the output of the regressor specific to the ground-truth class for each positive box.

The distillation loss $\mathcal{L}^{p}_{dist}$ is composed of two elements, one for the distillation of the classification scores $\mathcal{L}^{p}_{dist\_cls}$ and another one, $\mathcal{L}^{p}_{dist\_reg}$, for the regression:
\begin{equation}
 \mathcal{L}^{p}_{dist} = \mathcal{L}^{p}_{dist\_cls} + \mathcal{L}^{p}_{dist\_reg}~~.
\end{equation}
Given a box, the goal of the distillation loss is to make the output of the whole-body network as close as possible to the output of the part expert $p$.
The classification component $\mathcal{L}^{p}_{dist\_cls}$ is a standard distillation loss between the predictions produced by the corresponding part expert and those estimated by the whole-body model for part $p$.
In other words, $\mathcal{L}^{p}_{dist\_cls}$ is the soft version of hard label loss $\mathcal{L}^{p}_{cls}$.
The regression component $\mathcal{L}^{p}_{dist\_reg}$ is a L1 loss between the pose predicted by the part expert and the one estimated by the whole-body model for the ground-truth class.
Note that the pseudo ground-truth pose is obtained by averaging all overlapping estimates made by the part expert. While $\mathcal{L}^{p}_{reg}$ is designed to enforce regression of this pseudo ground-truth pose, $\mathcal{L}^{p}_{dist\_reg}$ favors regression of the exact same pose predicted by the part expert for a given box.

In practice, proposals generated by the RPNs of  part experts and  whole-body model are different
but computing distillation losses requires some proposals to coincide. 
At training, we thus augment the proposals of the whole-body model with positive boxes from the part experts to compute these losses.
In summary, given a training image, we: 
(a) run each part expert, keeping the positive boxes with classification scores and regression outputs, 
(b) run the whole-body model, adding the positive boxes from the experts to the list of proposals. 
Losses based on pseudo ground-truths are then averaged over all boxes while distillation losses are averaged only over positive boxes from the part experts.

\subsection{Training details}

\noindent \textbf{Data.} We train our body expert on the same combination of the MPII~\cite{mpii}, COCO~\cite{coco}, LSP~\cite{lsp}, LSPE~\cite{lspe}, Human3.6M~\cite{h36}  and Surreal~\cite{surreal} datasets  augmented with pseudo 3D ground-truth annotations as in~\cite{lcrnet++}. We applied random horizontal flips while training for 50 epochs. We train our hand expert on the RenderedHand (RH) dataset~\cite{renderedhand} for 100 epochs, with color jittering, random horizontal flipping and perspective transforms. $K_H=5$ anchor poses are obtained by clustering the 3D poses of  right  and flipped left hands from the training set. Finally, we train the face expert for 50 epochs on the Menpo dataset~\cite{menpo3d} with random horizontal flips and color jittering during training.

\noindent \textbf{Implementation.}
We implement DOPE in Pytorch~\cite{pytorch}, following the Faster R-CNN implementation from Torchvision.  We consider a ResNet50 backbone~\cite{resnet}.
We train it for 50 epochs, using the union of the datasets of each part expert, simply doubling the RH dataset used for hands as the number of images is significantly lower than for the other parts. The same data augmentation strategy used for training each part expert is employed for the whole-body network. We use Stochastic Gradient Descent (SGD) with a momentum of 0.9, a weight decay of $0.0001$ and an initial learning rate of 0.02, which is divided by 10 after 30 and 45 epochs. 
All images are resized such that the smallest image dimension is 800 pixels during training and testing and 1000 proposals are kept at test time.

\noindent \textbf{Runtime.}
DOPE runs at 100ms on a single NVIDIA T4 GPU.
When reducing the smallest image size to 400px and the number of box proposals to 50, and using half precision, it runs at 28 ms per image, \ie, in real-time at 35 fps, with a 2-3\% decrease of performance.
For comparison, each of our experts runs at a similar framerate as our whole-body model since only the last layers change. Optimization-based 3D whole-body estimation methods~\cite{smplx,totalcapture} take 
up to a minute to process each person.

\section{Experiments}
\label{sec:xp}

Given that there is no dataset to evaluate whole-body 3D pose estimation in the wild, we evaluate our method on each task separately.
After presenting datasets and metrics (Section~\ref{sub:datametric}), we compare the performance of our whole-body model to the experts (Section~\ref{sub:res1}) and to the state of the art (Section~\ref{sub:res2}).

\subsection{Evaluation datasets and metrics}
\label{sub:datametric}

\noindent \textbf{MPII for 2D body pose estimation.}
As in~\cite{lcrnet++}, we remove 1000 images from the MPII~\cite{mpii} training set and use them to evaluate our 2D body pose estimation results.
We follow the standard evaluation protocol and report the PCKh@0.5 which is the percentage of correct keypoints with a keypoint being considered as correctly predicted if the error is smaller than half the size of the head.

\noindent \textbf{MuPoTs for 3D body pose estimation.}
MuPoTs-3D~\cite{xnect} (Multi-person Pose estimation Test Set in 3D) is composed of more than 8,000 frames from 20 real-world scenes with up to three subjects. The ground-truth 3D poses, obtained using a multi-view MoCap system, have a slightly different format than the one estimated by our body expert and whole-body model.  To better fit their 14-joint skeleton model, we modified the regression layer of our networks to output 14 keypoints instead of 13 while freezing the rest of the network. 
We finetuned this last layer only on the MuCo-3DHP dataset~\cite{xnect}, the standard training set when testing on MuPoTs.
We report the 3D-PCK, \ie, the percentage of joint predictions with less than 15cm error, per sequence, and averaged over the subjects for which ground truth is available.

\noindent \textbf{RenderedHand for 3D hand pose estimation.}
RenderedHand (RH) test set~\cite{renderedhand} consists of 2,728 images showing the hands of a single person.
We report the standard AUC (Area Under the Curve) metric when plotting the 3D-PCK  after normalizing the scale and relative translation between the ground-truth and the prediction. Note that while state-of-the-art methods evaluate hand pose estimation given ground-truth crops around the hands, we instead perform an automatic detection but miss around 2\% of the hands.

\noindent \textbf{Menpo for facial landmark estimation.}
We report results for facial landmark evaluation using the standard 3D-aware 2D metric~\cite{menpo} on the 30 videos from the test set of the ICCV'17 challenge~\cite{menpo3d}.
Given a ground truth-matrix ${\bm{s}} \in \mathcal{M}_{N, 2}(\mathbb{R})$ representing the 2D coordinates in the image of the $N=84$ landmarks of a face, and a facial landmark prediction $\hat{\bm{s}} \in \mathcal{M}_{N, 2}(\mathbb{R})$, this 2D normalized point-to-point RMS error is defined as:
\begin{equation}
\epsilon({\bm{s}}, \hat{\bm{s}}) = \cfrac{ \|{\bm{s}} - \hat{\bm{s}} \|_2}{\sqrt{N} d_{scale}}~~,
\end{equation}
where $d_{scale}$ is the length of the diagonal of the minimal 2D bounding box of  ${\bm{s}}$.

\begin{table}
 \centering
 \caption{Comparison between our part experts and our whole-body model}
 \begin{tabular}{lcccc}
 \toprule
  \multirow{2}{*}{ }              & MPII         & MuPoTs       & RH test & Menpo \\
                                      & (PCKh@0.5)   & (PCK3D)      & (AUC)   & (AUC) \\
  \midrule
  body expert                         & \textbf{89.6}& 66.8 & -     & -     \\
  hand expert                         & -            & -        & \textbf{87.1} & -     \\
  face expert                         & -            & -        & -     & 73.9 \\
  \midrule
  whole-body trained on gt                       & 88.3         & 66.6 & 81.1 & 61.7 \\
  \midrule
  DOPE without $\mathcal{L}^p_{dist}$   & 88.3         & 66.4 & 83.5 & \textbf{75.2} \\
  DOPE with $\mathcal{L}^p_{dist}$      & 88.8         & \textbf{67.2} & 84.9 & 75.0 \\
  \bottomrule
 \end{tabular}
 \label{tab:res}
\end{table}

\subsection{Comparison to the experts}
\label{sub:res1}

Table~\ref{tab:res} presents a comparison of the performances obtained by the part experts and our DOPE model, for body, hand and face pose estimation tasks.

We first compare the part experts to a baseline where our whole-body network is trained on the partial ground-truth available for each dataset, \eg only body annotations are available on images from body datasets, \etc
The performance degrades quite significantly compared to those of the hand and face experts (-6\% AUC for hand on the RH dataset and -12\% for face landmarks).
This is explained by a lower detection rate of the detector due to the fact that, for instance, unannotated faces present in the body datasets are considered as negatives during training.
The performance of this model on body pose estimation is quite similar to the one of the body expert: as bodies are not observed too much in hand and face datasets, there are almost no missing body annotations.

We then compare the experts to a first version of our DOPE model without the distillation loss $\mathcal{L}^p_{dist}$.
The performance on body pose estimation remains similar but, for hands and faces, a significant gain is obtained, in particular for faces, 
where the whole-body network performs even better than the expert.
This might be explained by the fact that the whole-body network is trained on a larger variety of data, including images from body and hands datasets with many additional faces.
In contrast, the hand component performs slightly lower than the expert. One hypothesis is that many hands in the body datasets are too small to be accurately estimated, leading to noisy pseudo ground-truth poses. However, the performance remains close to that of the hand expert.

\begin{figure}[t]
 \centering
 \rotatebox{90}{~~~~~~~body} \hfill
 \includegraphics[width=0.315\linewidth]{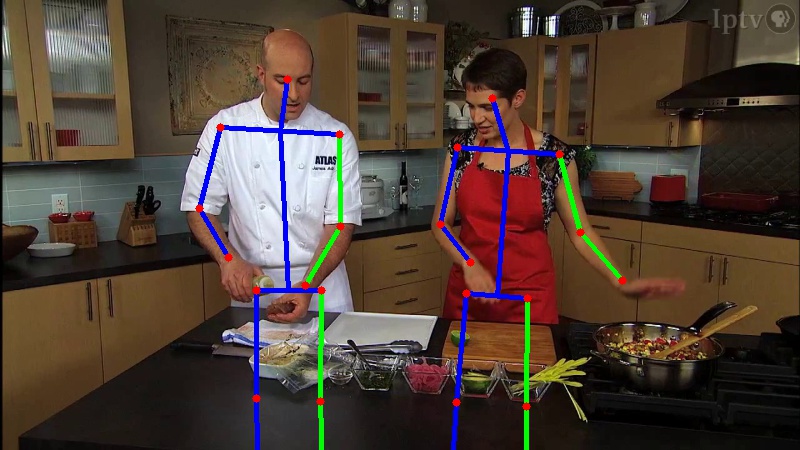} \hfill
 \includegraphics[width=0.315\linewidth]{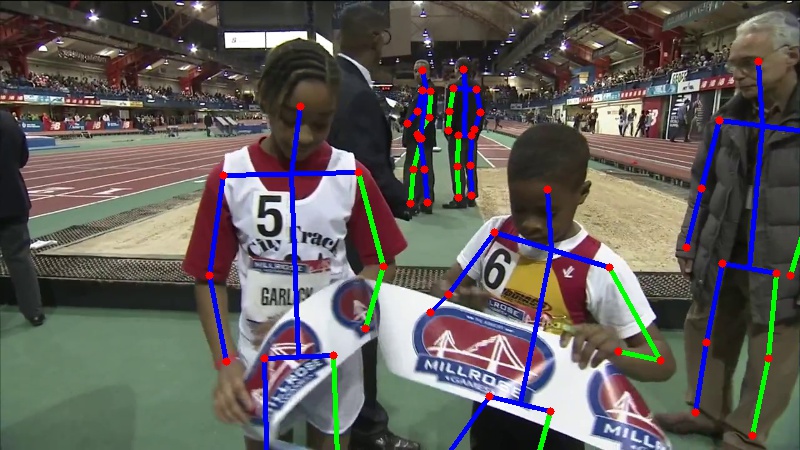} \hfill
 \includegraphics[width=0.315\linewidth]{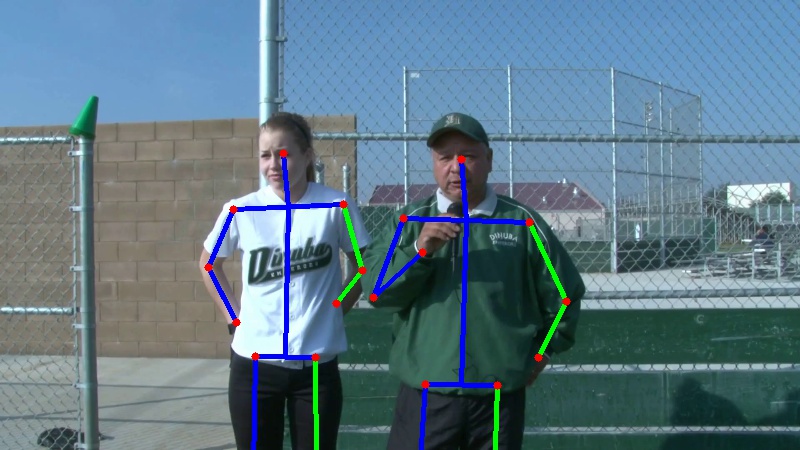} \\[0.1cm]
 \rotatebox{90}{~~~~~~hands} \hfill
 \includegraphics[width=0.315\linewidth]{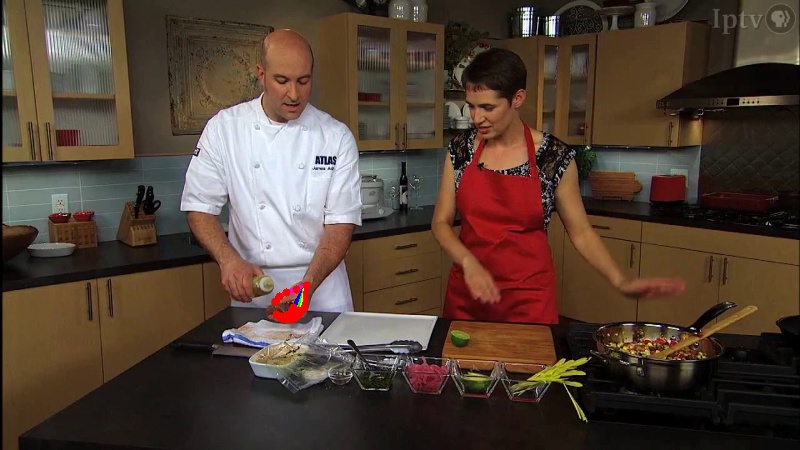} \hfill
 \includegraphics[width=0.315\linewidth]{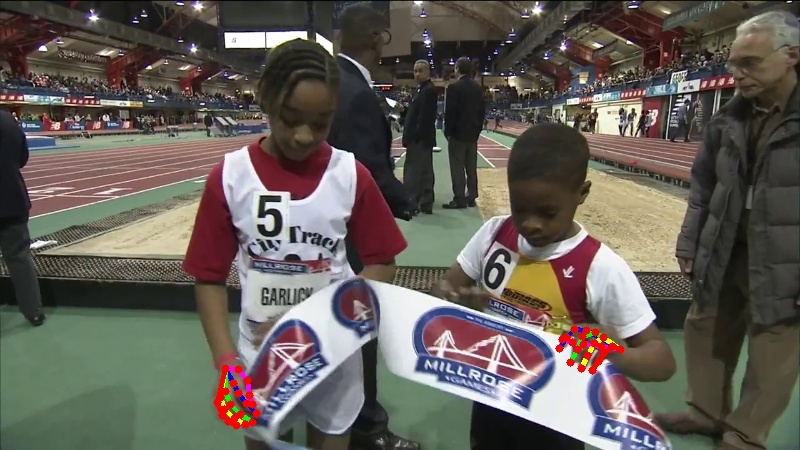} \hfill
 \includegraphics[width=0.315\linewidth]{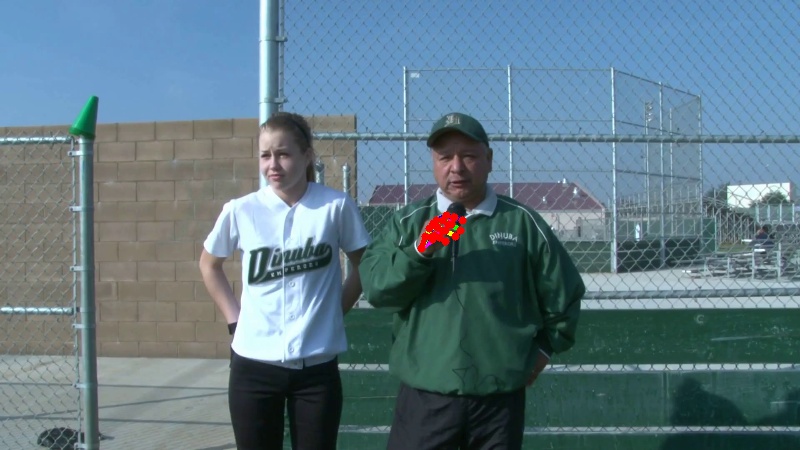} \\[0.1cm]
 \rotatebox{90}{~~~~~~~face} \hfill
 \includegraphics[width=0.315\linewidth]{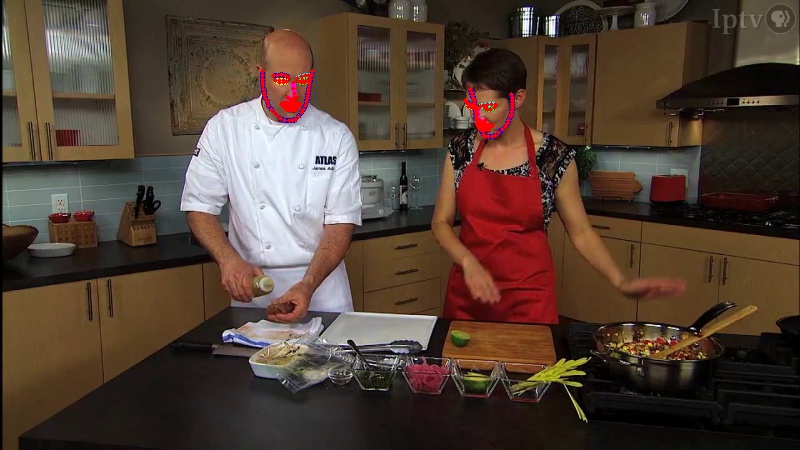} \hfill
 \includegraphics[width=0.315\linewidth]{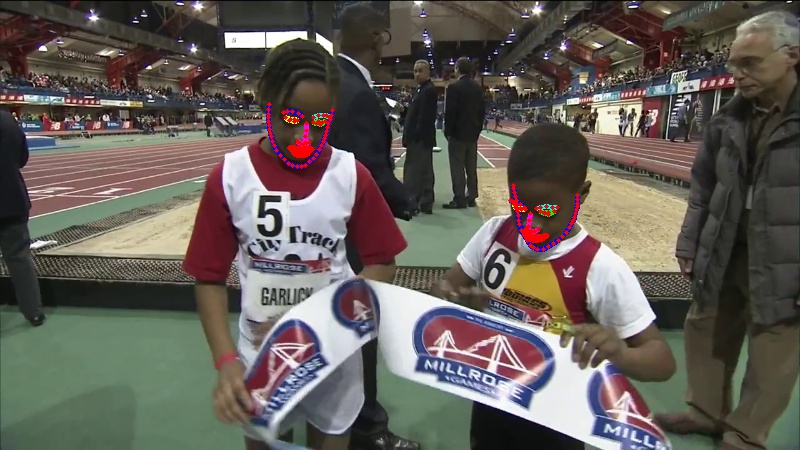} \hfill
 \includegraphics[width=0.315\linewidth]{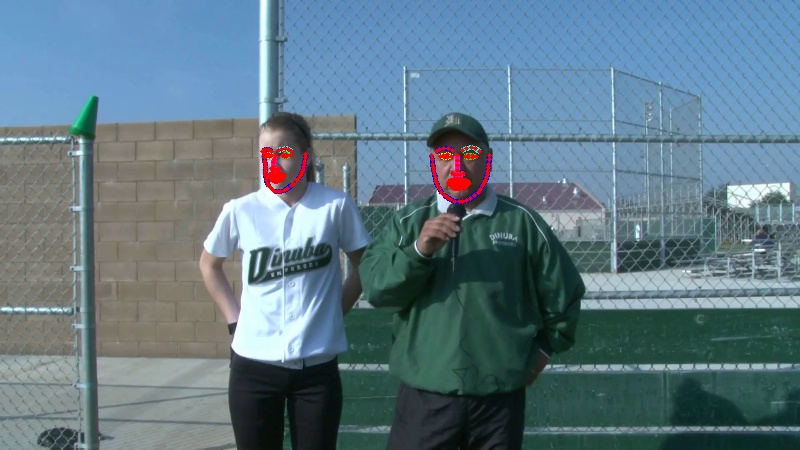} \\[0.1cm]
 \rotatebox{90}{~\textbf{DOPE (2D)}} \hfill
 \includegraphics[width=0.315\linewidth]{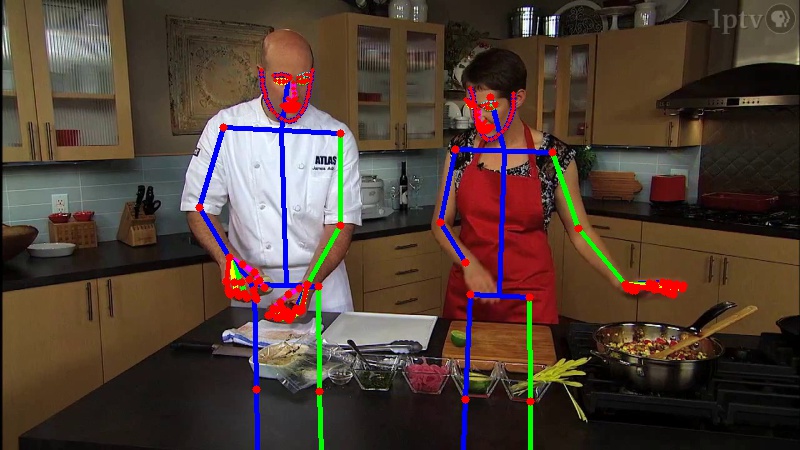} \hfill
 \includegraphics[width=0.315\linewidth]{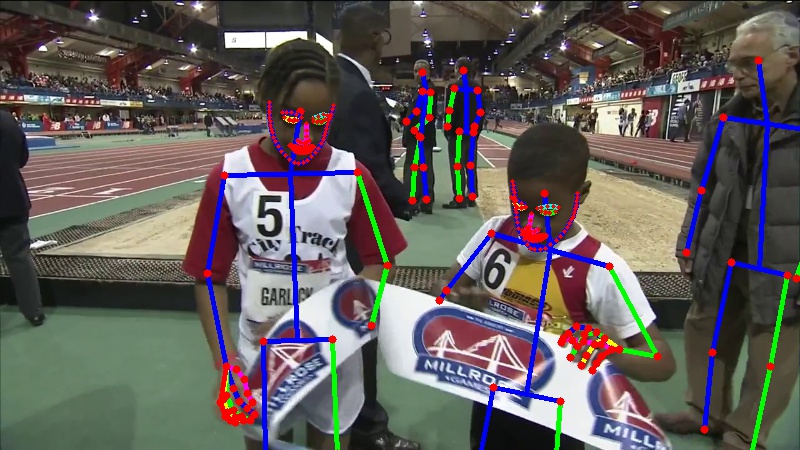} \hfill
 \includegraphics[width=0.315\linewidth]{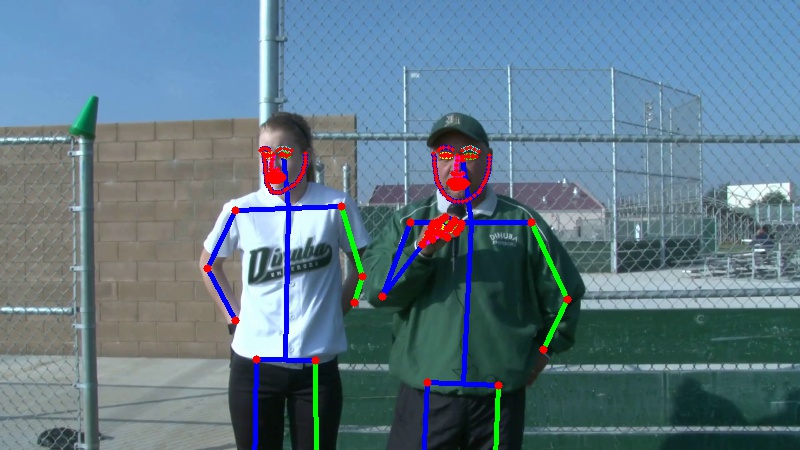} \\[-0.1cm]
 \rotatebox{90}{~~~~~~~~\textbf{DOPE (3D)}} \hfill
 \includegraphics[trim=200 80 200 110,clip, width=0.315\linewidth]{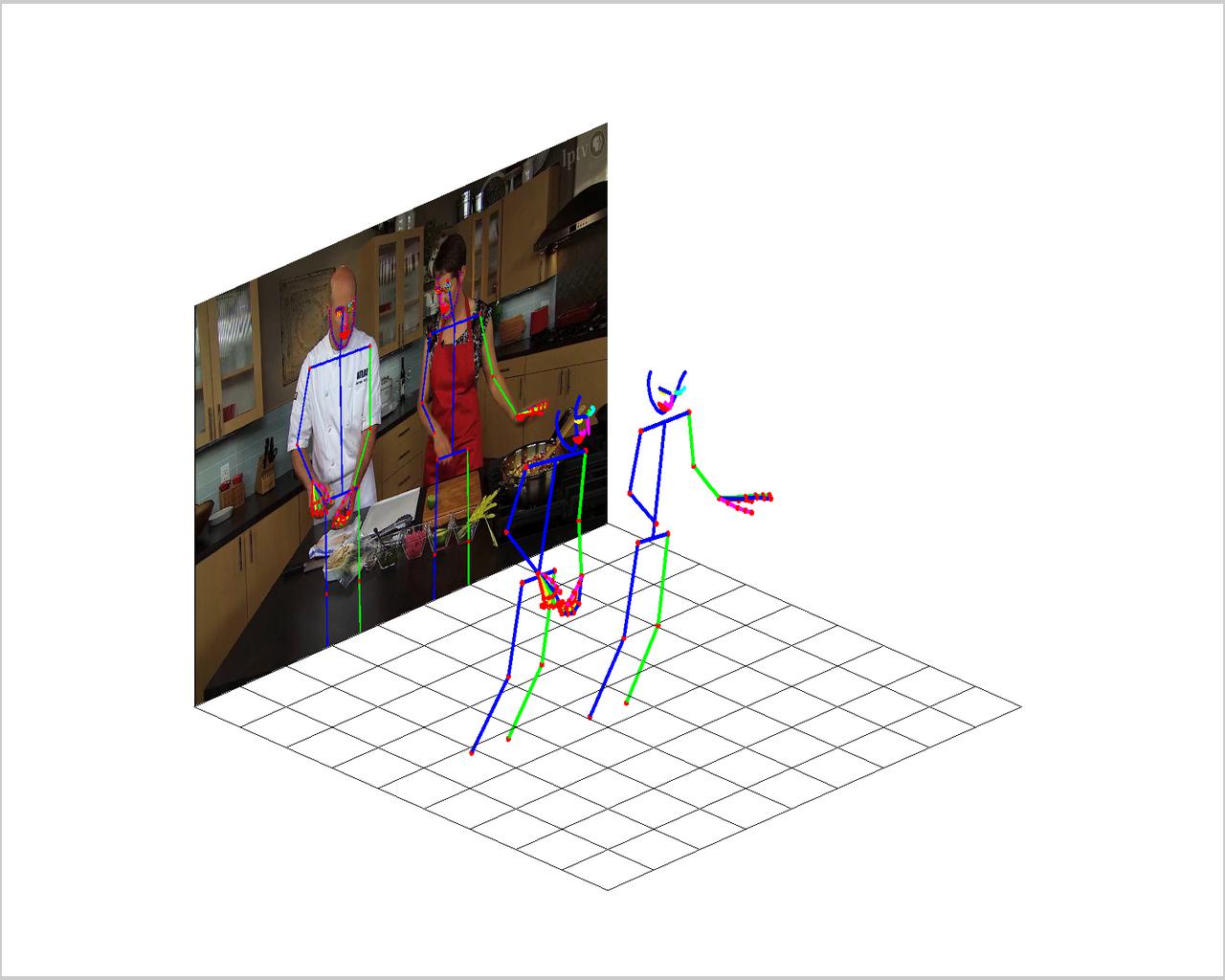} \hfill
 \includegraphics[trim=240 80 200 140,clip, width=0.315\linewidth]{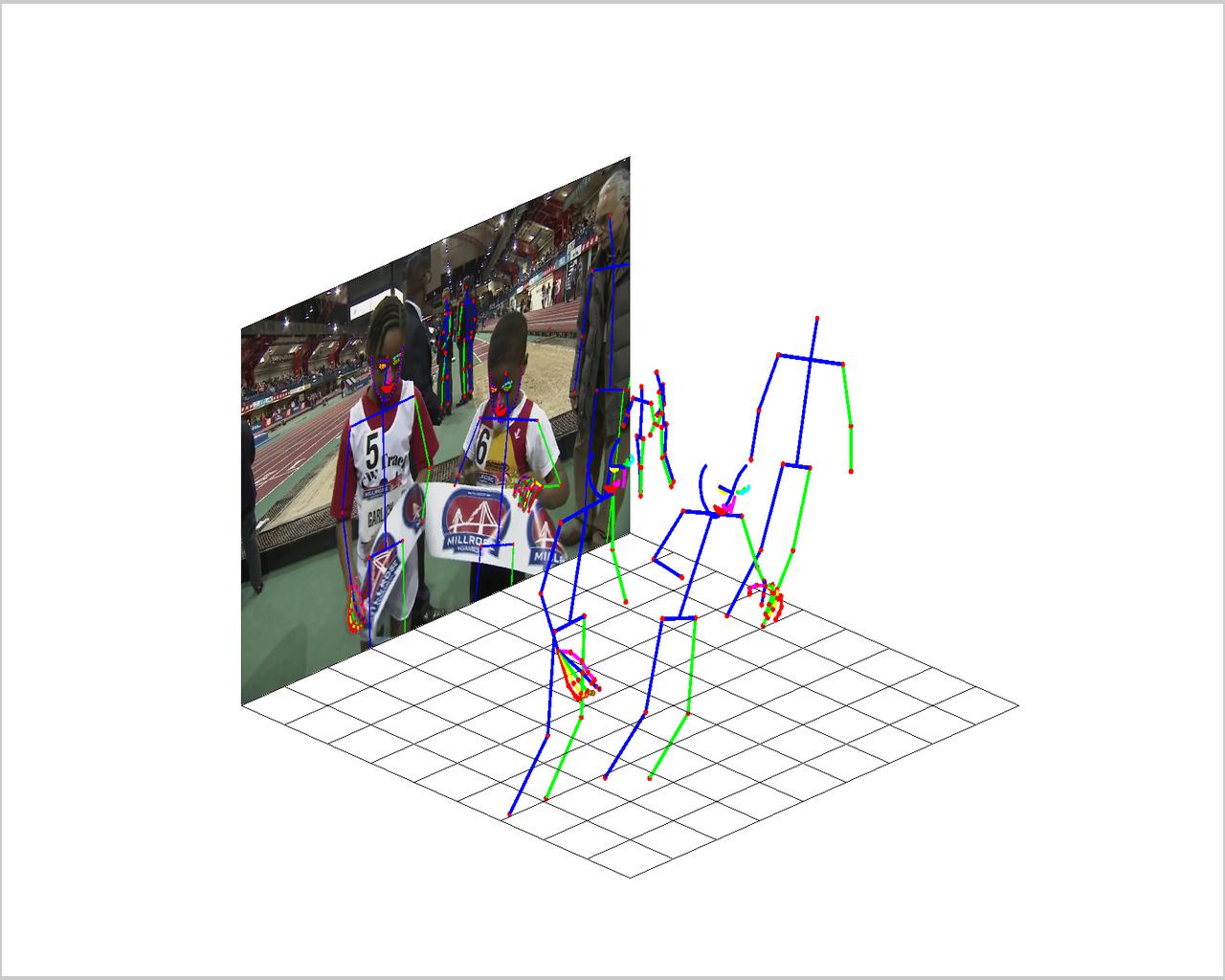} \hfill
 \includegraphics[trim=240 80 200 140,clip, width=0.315\linewidth]{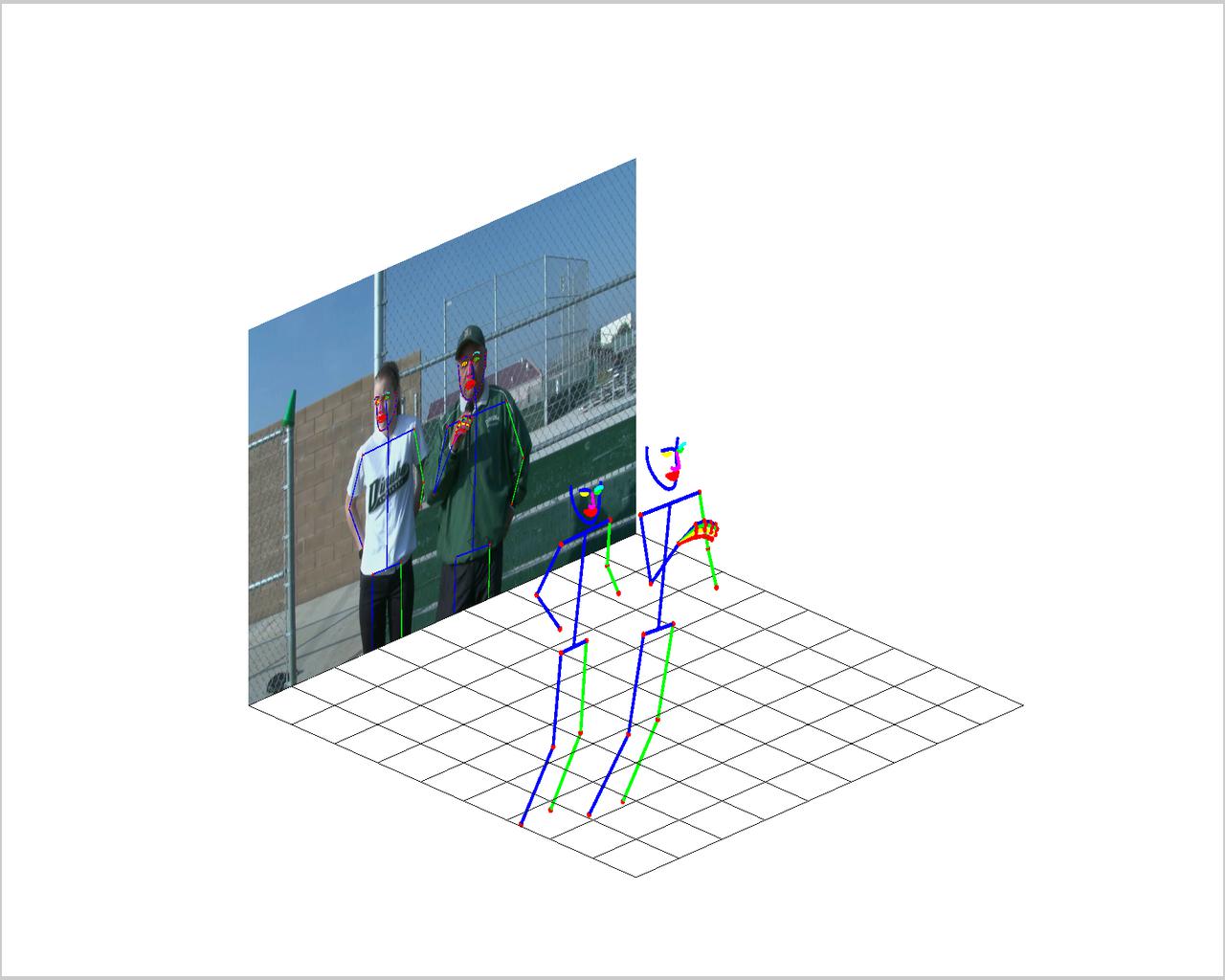} \\[-0.35cm]
 \caption{Each column shows an example with the results of the 3 experts on the first three rows (we show only the 2D for clarity). The last two rows show the results obtained by our proposed DOPE approach in 2D and in 3D respectively.}
 \vspace{-0.6cm}
 \label{fig:resexamples}
\end{figure}

With the addition of the distillation loss, the accuracy increases for hand pose estimation (+1.4\%) and slightly for body pose estimation (+0.5\% on MPII, +0.8\% on MuPoTs), bringing the performance of the whole-body network even closer to the experts' results.
Sometimes, DOPE even outperforms the part expert as observed on MuPoTs for multi-person 3D pose estimation or on Menpo for facial landmark detection.
Figure~\ref{fig:resexamples} presents some qualitative results for the part experts and our proposed DOPE model trained with distillation loss. Two additional examples of our model's results can be found in Figure~\ref{fig:res}. DOPE produces high-quality whole-body detections that include bodies, hands and faces.
In the example on the left in Figure~\ref{fig:resexamples}, our whole-body network correctly detects and estimates the pose of three hands, misdetecting only the lady's right hand.  By contrast, the hand expert only finds one hand in this image. 
Note that our method is holistic for each part: if a part is sufficiently visible in the image, a prediction is made for every keypoint of the part despite partial occlusions or truncations, as shown for the bodies in this same example. However, if a part is not visible, no prediction is made. This is the case for the occluded hands in the middle and right examples in Figure~\ref{fig:resexamples}.
Overall, these examples illustrate that our method can be applied in the wild, including scenes with multiple interacting people, varied background, severe occlusions or truncations.

\begin{figure}
  \begin{tabular}{cc}
   \hspace{2mm}
  \begin{minipage}[c]{0.45\linewidth}
    \hspace{2mm}
  \includegraphics[width=\linewidth]{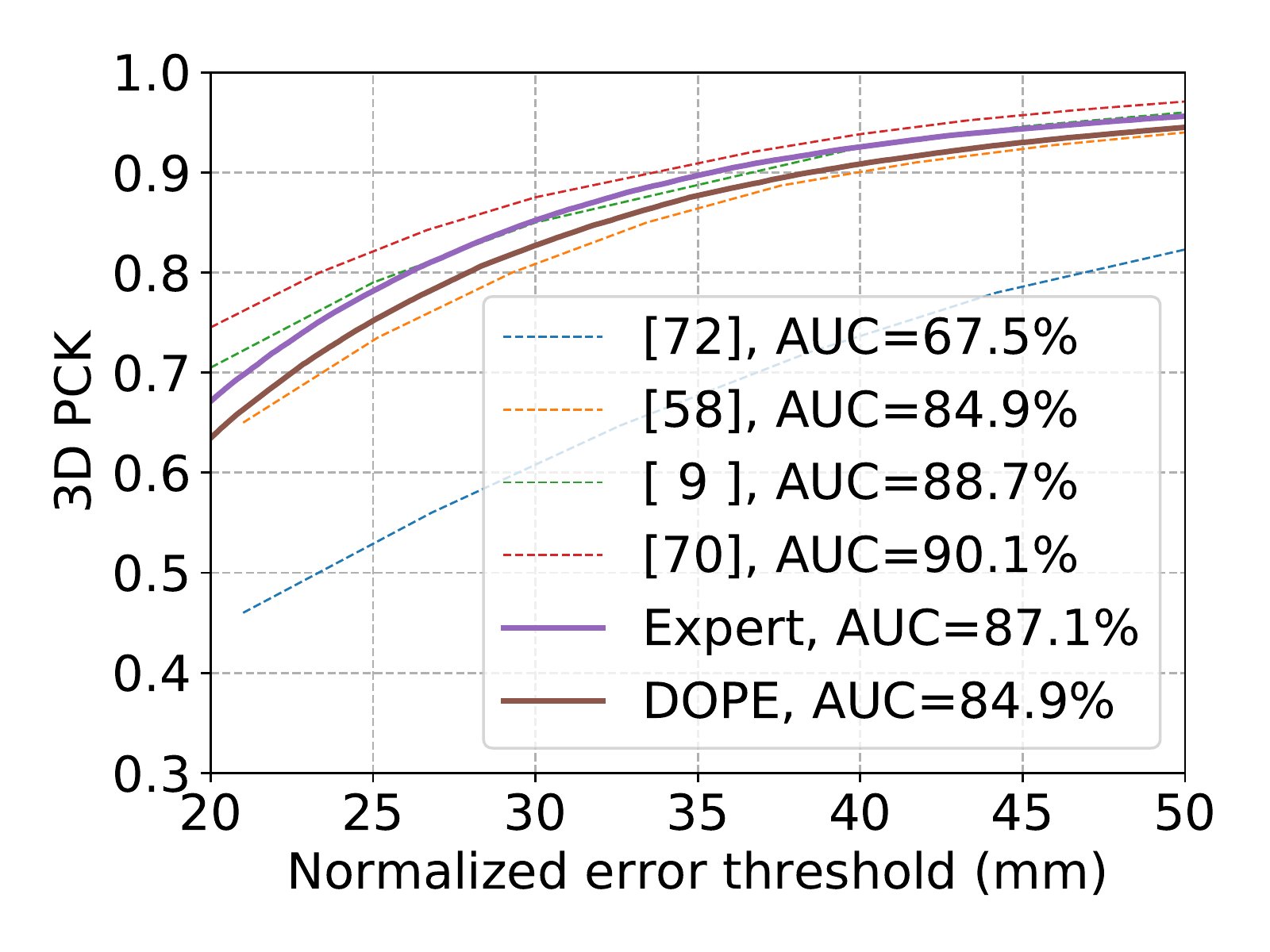}
  \end{minipage}
  &
  \begin{minipage}[c]{0.45\linewidth}
  \hspace{3mm}
  \includegraphics[width=\linewidth]{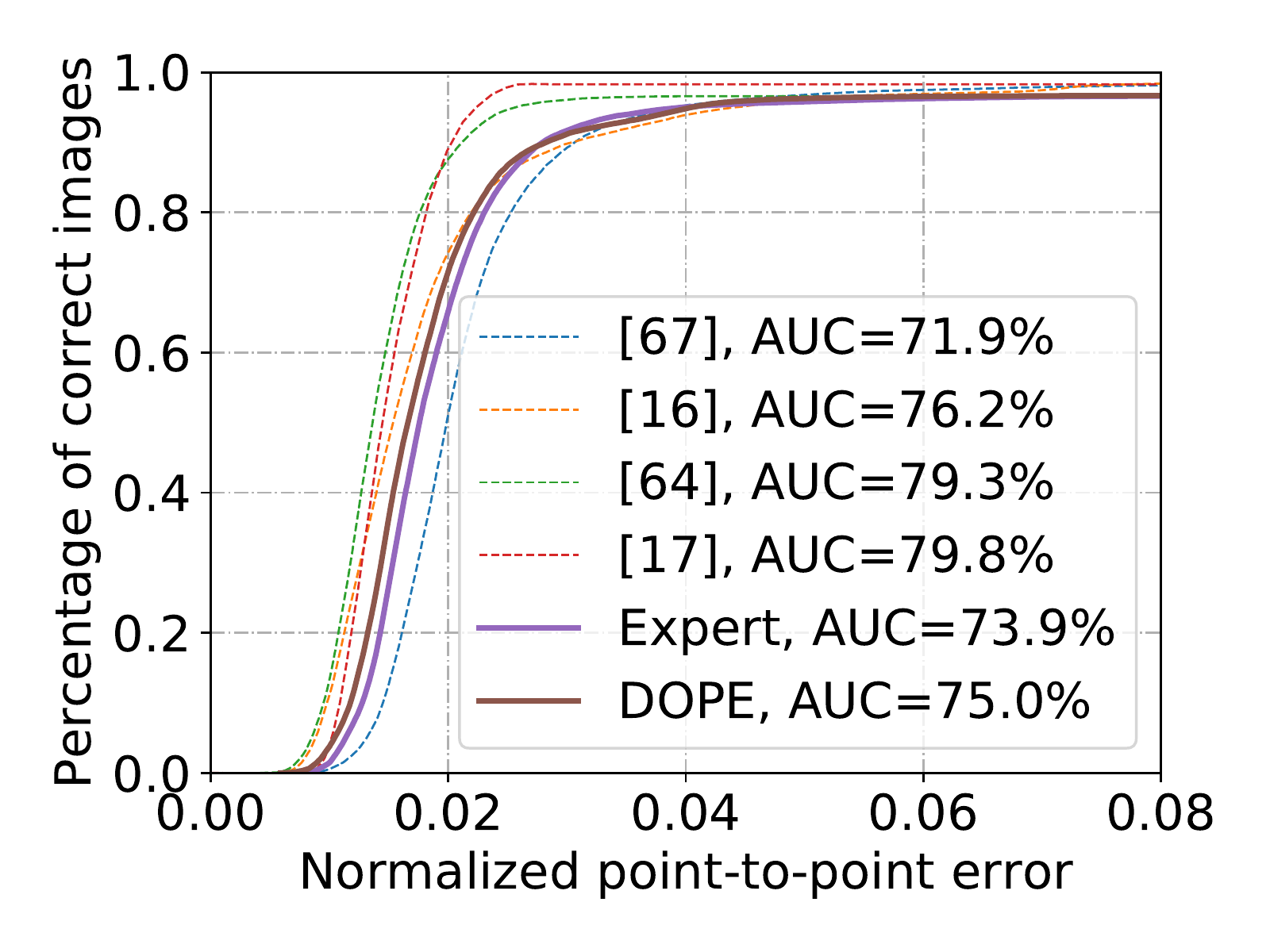}
  \end{minipage}

    \\[-0.2cm]
  (a) & (b)

  \\[0.2cm]

  \begin{minipage}[c]{0.25\linewidth}
  \resizebox{1.1\linewidth}{!}{
  \begin{tabular}{cc}
  \toprule
   method & PCK3D \\
   \midrule
   VNect~\cite{vnect} & 65.0 \\
   XNect~\cite{xnect} & 70.4 \\
   LCRNet++~\cite{lcrnet++} & \textbf{70.6} \\
   \midrule
   body expert & 66.8 \\
   \textbf{DOPE} & 67.2 \\
   \bottomrule
  \end{tabular}
  }
  \end{minipage}
  
  & 
  \begin{minipage}[c]{0.4\linewidth}
  \resizebox{1.1\linewidth}{!}{
  \begin{tabular}{ccc}
  \toprule
   method & ~~RH (PCK2D)~~ & ~~Menpo (PCK2D)~~\\
   \midrule
   Experts & \textbf{78.1} & 94.5 \\
     \textbf{DOPE}  &  71.0 & \textbf{94.9} \\ 
   \midrule
\cite{openpose}  & 36.1 & 88.0 \\
\cite{wholebody} & 23.5 & 71.5 \\
   \bottomrule
  \end{tabular}
  }
  \end{minipage}
   \\
  (c) & (d)
      \end{tabular}
    \vspace{-0.3cm}
  \caption{Comparison to the state of the art: (a) PCK3D on RH for varying error threshold (hand). (b) Percentage of images with correct face detections for varying 3DA-2D thresholds on Menpo (face). (c) PCK3D on MuPoTs (body). (d) 2D PCK at a threshold of 10\% of the tight bounding box's largest size on RH (hand) and 5\% on Menpo (face). The higher the better.}
  \label{fig:sota}  
  \vspace{-0.5cm}
\end{figure}

\begin{figure}[htb]
 \centering
 \rotatebox{90}{~~~~~~image} \hfill
 \includegraphics[width=0.315\linewidth]{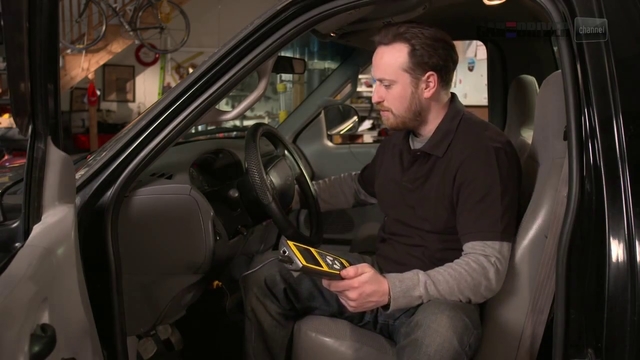} \hfill
 \includegraphics[width=0.315\linewidth]{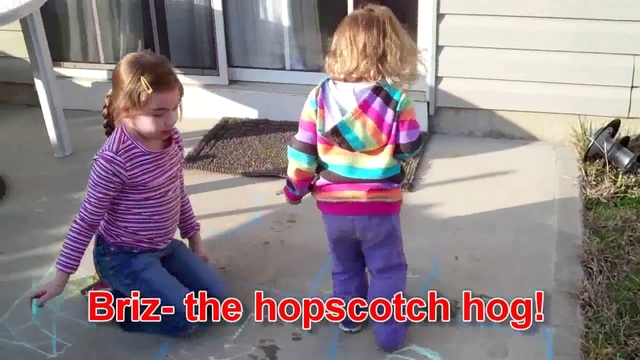} \hfill
 \includegraphics[width=0.315\linewidth]{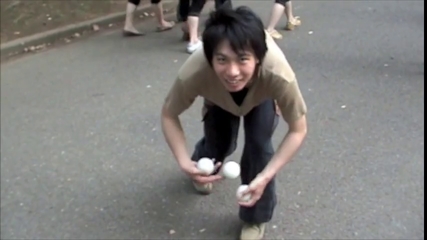} \\
 \rotatebox{90}{~~~~MTC~\cite{totalcapture}} \hfill
 \includegraphics[width=0.315\linewidth]{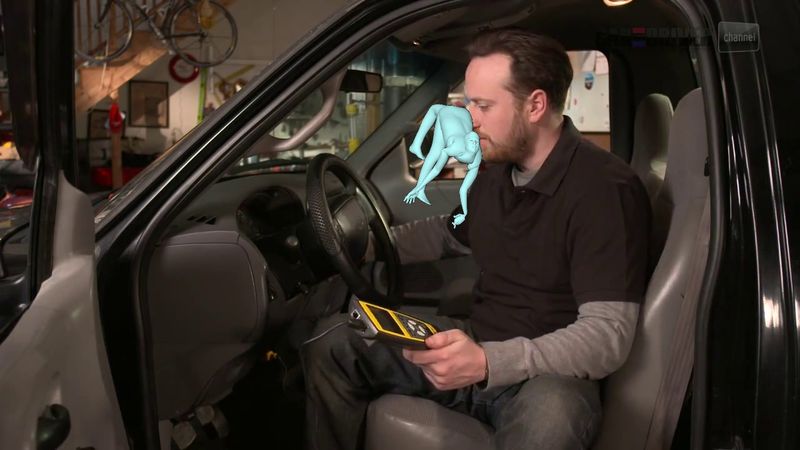} \hfill
 \includegraphics[width=0.315\linewidth]{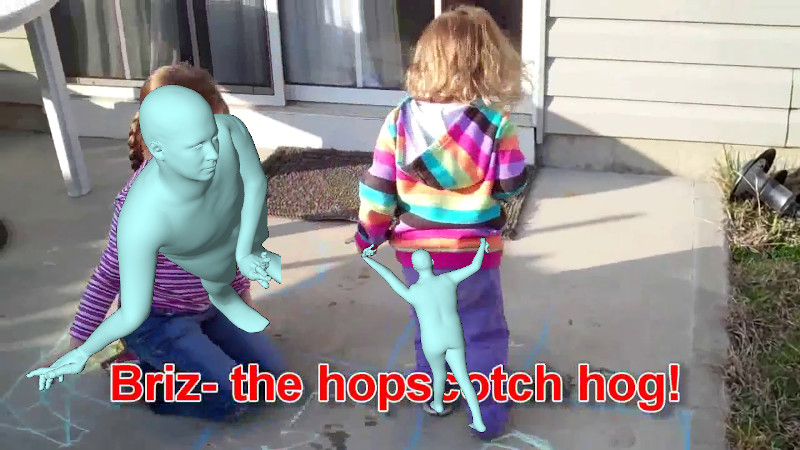} \hfill
 \includegraphics[width=0.315\linewidth]{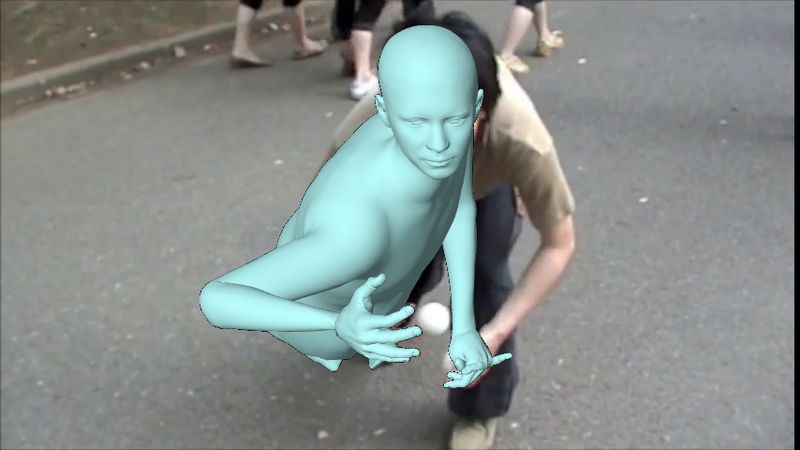} \\
 \rotatebox{90}{video-MTC~\cite{totalcapture}} \hfill
 \includegraphics[width=0.315\linewidth]{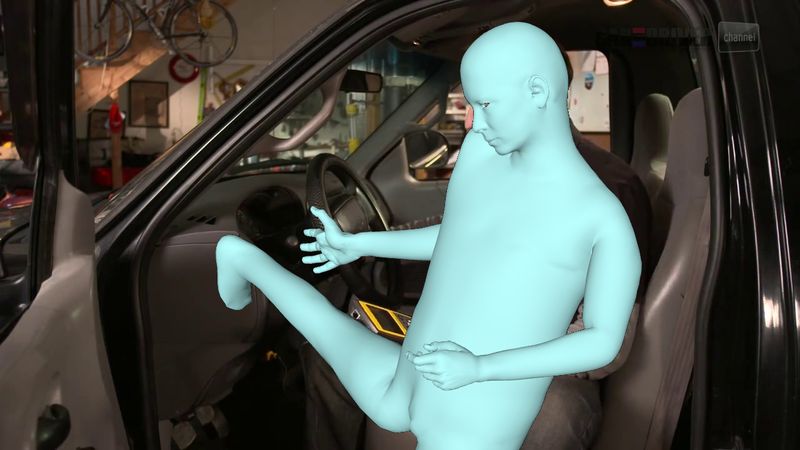} \hfill
 \includegraphics[width=0.315\linewidth]{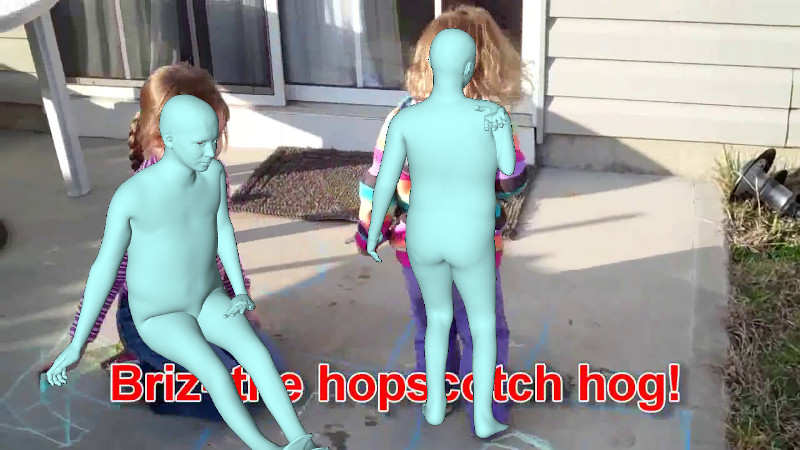} \hfill
 \includegraphics[width=0.315\linewidth]{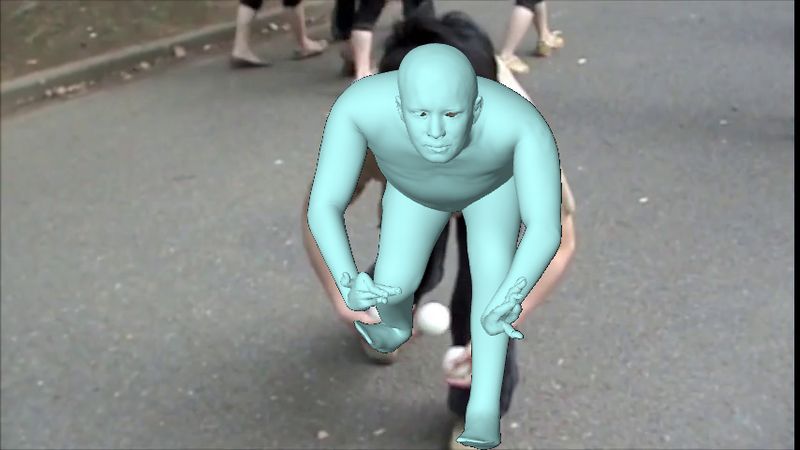} \\
 \rotatebox{90}{SMPLify-X~\cite{smplx}} \hfill
 \includegraphics[width=0.315\linewidth]{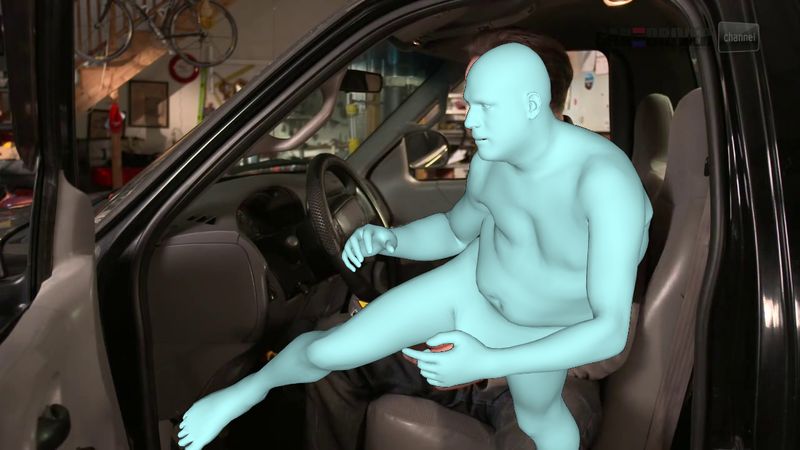} \hfill
 \includegraphics[width=0.315\linewidth]{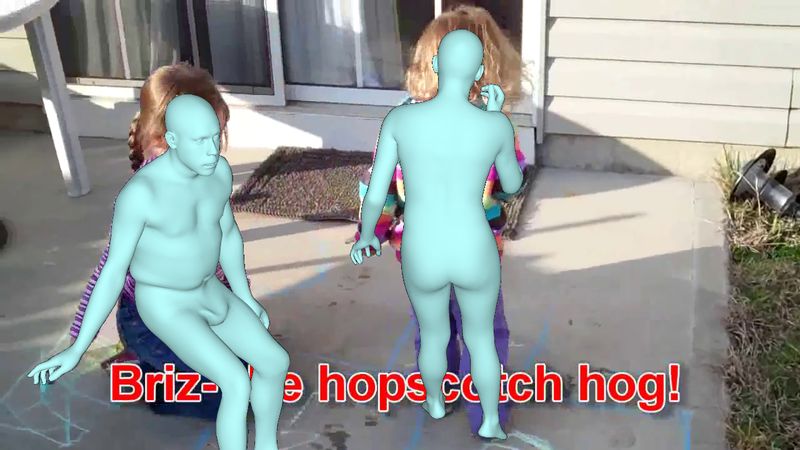} \hfill
 \includegraphics[width=0.315\linewidth]{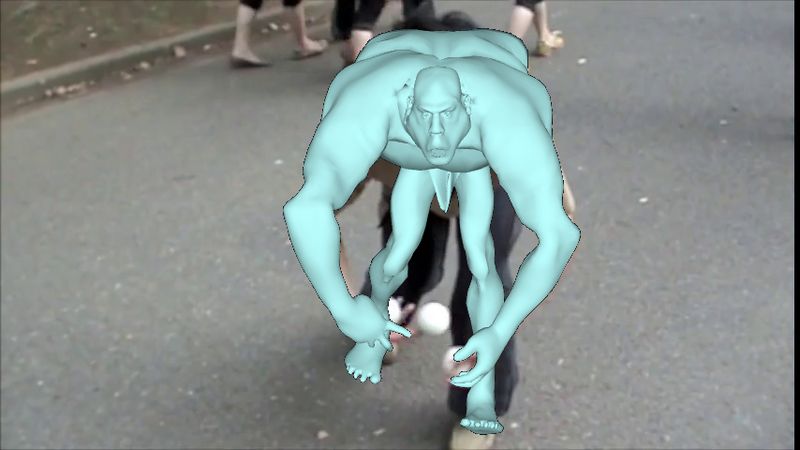} \\[0.0cm]
 \rotatebox{90}{~~~~~~~~~\textbf{DOPE}} \hfill
 \includegraphics[trim=200 80 200 110,clip,width=0.315\linewidth]{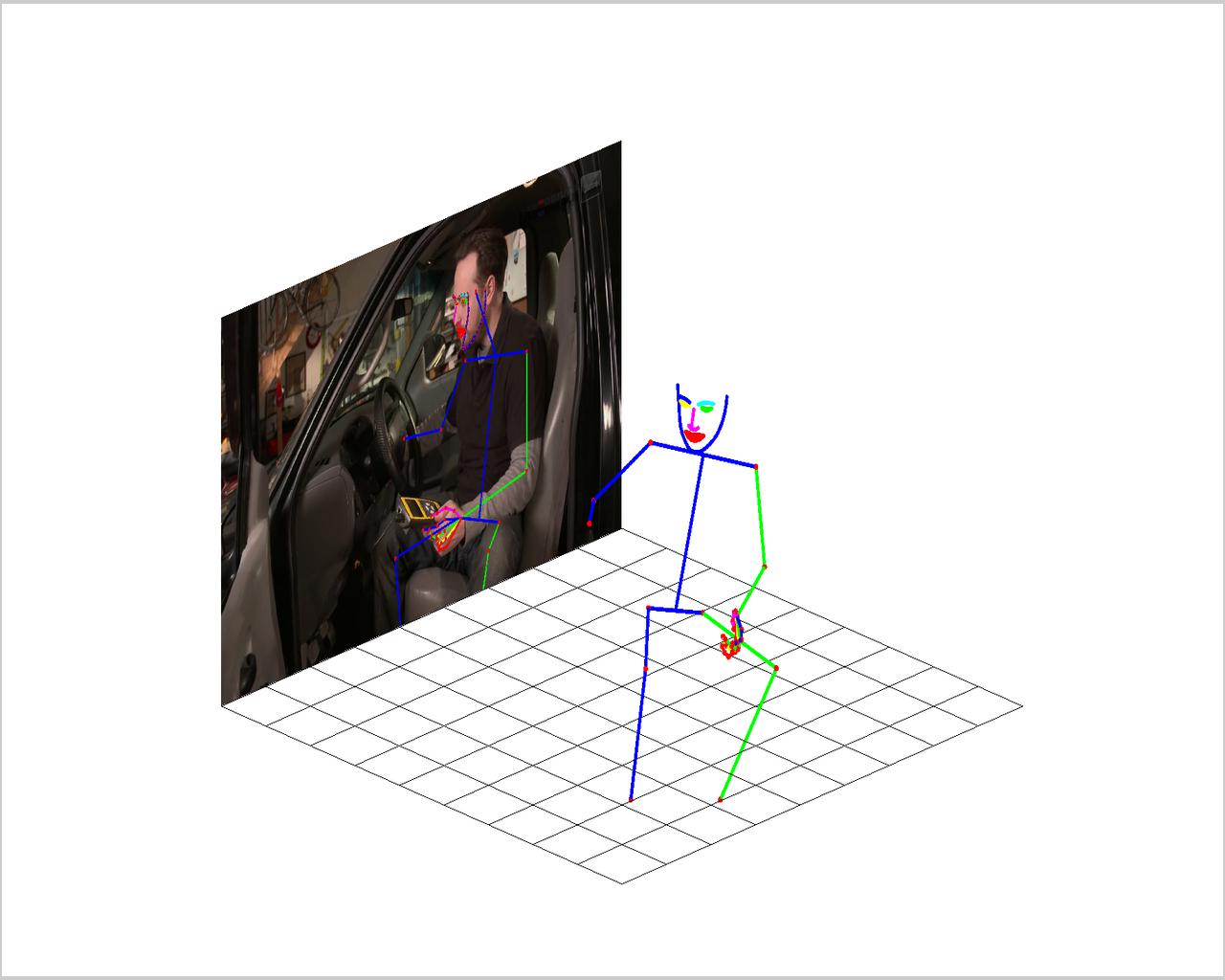} \hfill
 \includegraphics[trim=200 80 200 110,clip,width=0.315\linewidth]{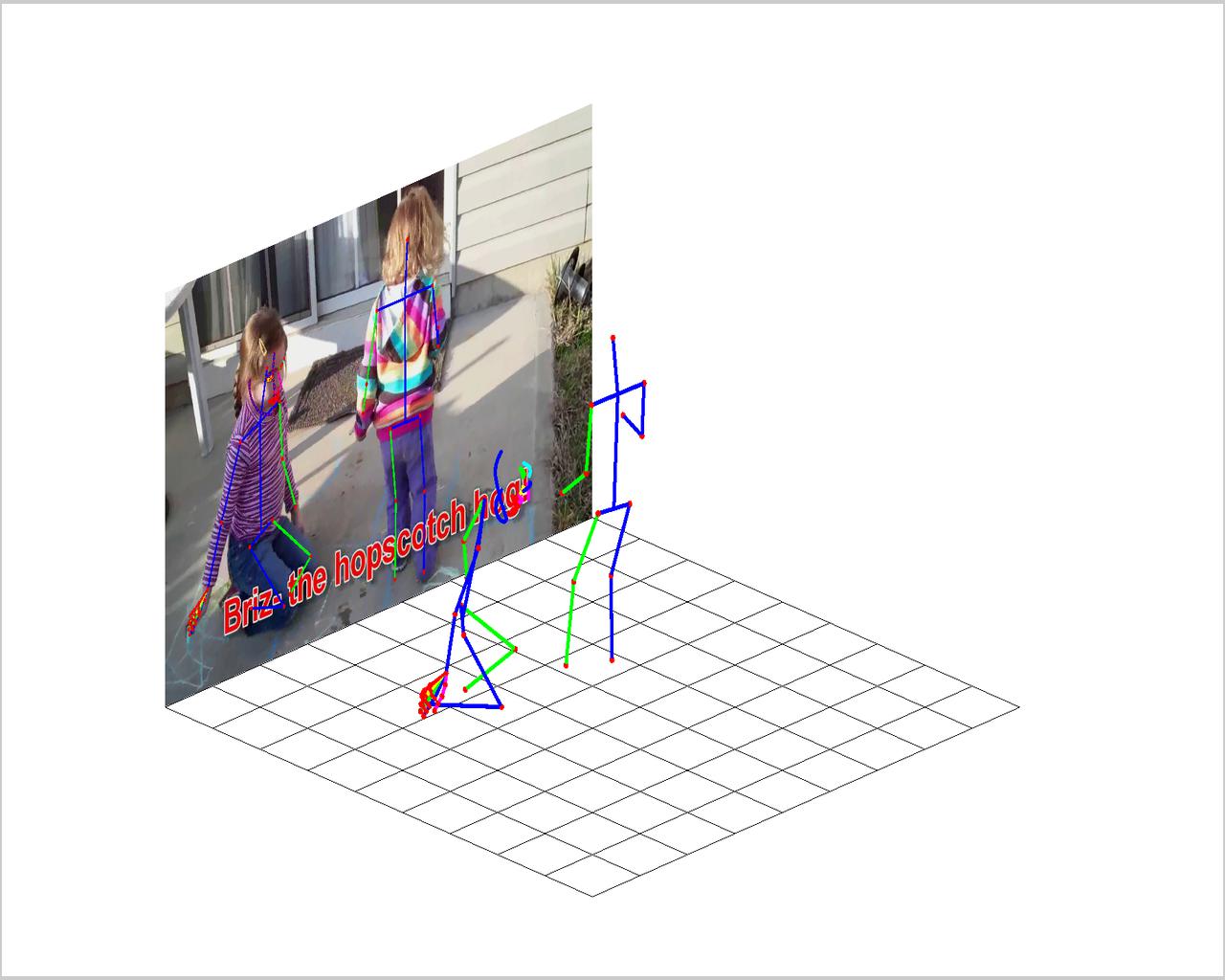} \hfill
 \includegraphics[trim=200 80 200 110,clip,width=0.315\linewidth]{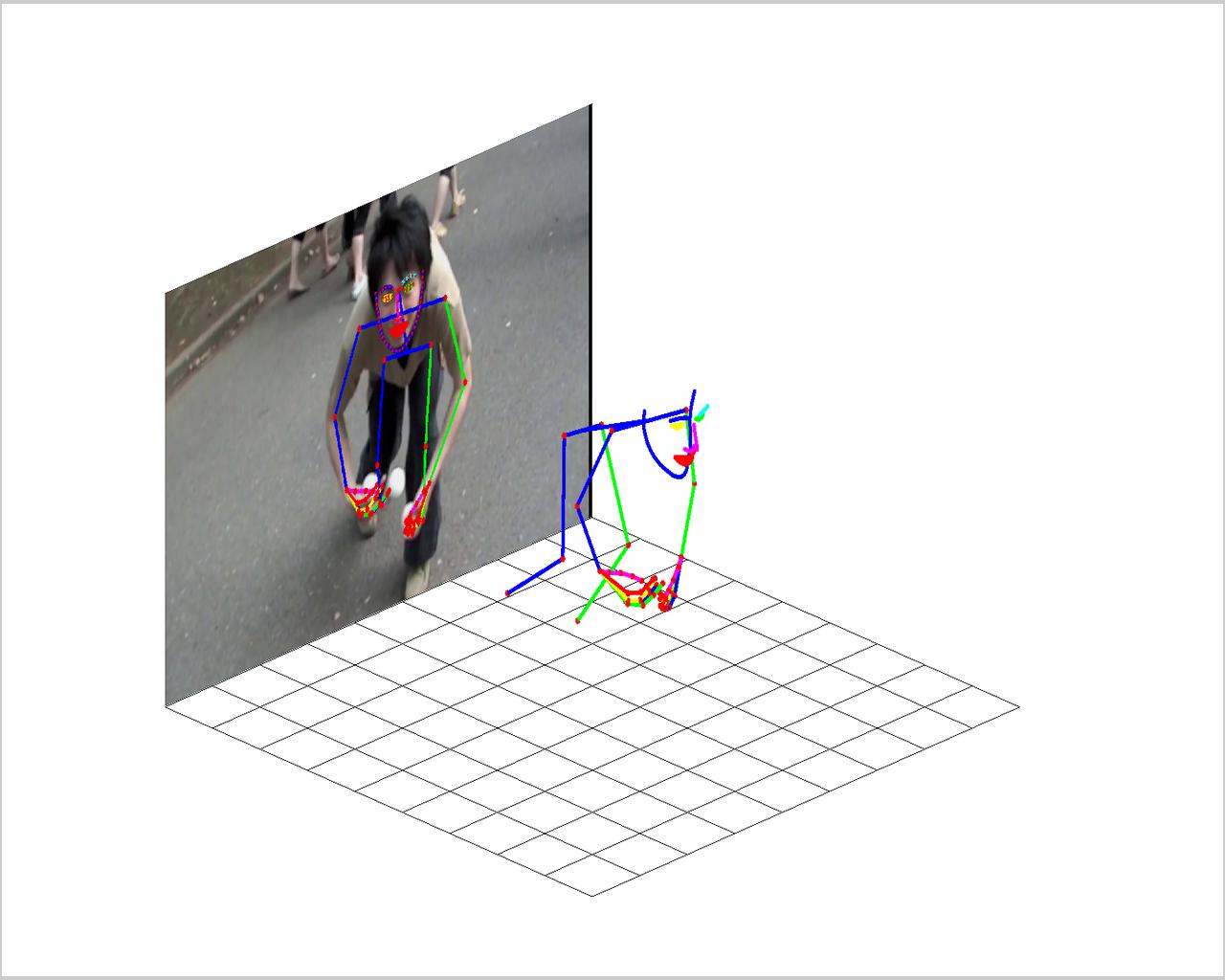} \\[-0.35cm]
 \caption{Three examples with from top to bottom the original image, the results from MTC~\cite{totalcapture} from static image or the video, SMPLify-X~\cite{smplx} and ours.}
 \label{fig:sotares}
\end{figure}
\subsection{Comparison to the state of the art}
\label{sub:res2}

\noindent \textbf{Comparison on individual tasks.}  In Figure~\ref{fig:sota}, we compare our DOPE approach to the state of the art for each individual task.
Note that our main goal is not to outperform the state of the art on each of these tasks but rather to unify 3 individual models into a single one while still achieving a competitive performance.
DOPE is among the top performing methods for all three tasks, \ie, hand  (a),  face (b) and body (c) 3D pose estimation, while being the first and only method to report on these three tasks together. Additionally, our detection network  tackles a more difficult task than most of our competitors who assume that a bounding box around the ground-truth hands~\cite{renderedhand, spurr2018, cai2018, zhang2019} or faces~\cite{xiong2017, zadeh2017, pix2face, menpo} is given at test time. We also compare with existing whole-body 2D pose estimation methods (d).

\noindent \textbf{Qualitative comparison to~\cite{totalcapture,smplx}.}
Since there is no dataset to numerically compare the performances of our learning-based approach in the wild against the optimization-based pipelines such as~\cite{totalcapture,smplx}, 
we show some qualitative examples in Figure~\ref{fig:sotares}.
We find that Monocular Total Capture~\cite{totalcapture} performs quite poorly on static images (second row), in particular due to occlusions.
It greatly benefits from temporal information when processing the sequences from which the images are extracted (third row).
However, there are still some errors, especially in case of occlusions (\eg legs in the left column image).
For~\cite{smplx} (fourth row), in the first example, OpenPose~\cite{openpose} does not estimate the 2D location of the feet that fall out of the field of view, impacting the optimization of the model's legs. In our case, the  pose of the legs is correctly estimated. The same phenomenon happens in the second example where a little girl is kneeling and the self-occlusions prevent her feet from being detected. Finally, in the third example, the optimization gets stuck in a local minimum while our estimation is more robust. 
In addition of its robustness in the wild, our learning-based approach is also about 1000 times faster than~\cite{smplx} which takes about a minute per person in an image.

\section{Conclusion}
\label{sec:conc}

We have proposed DOPE, the first learning-based method to detect and estimate whole-body 3D human poses in the wild, including body, hand and face 2D-3D keypoints.
We tackled the lack of training data for this task by leveraging distillation from part experts to our whole-body network.
Our experiments validated this approach showing performances close to the part experts' results.

Our method allows training a network on a more diverse set of in-the-wild images, potentially without any pose annotations. In future work, we will investigate if our model could benefit from additional unlabeled training data.

\clearpage

%
%
\bibliographystyle{splncs04}
\bibliography{biblio}

\end{document}